\documentclass[10pt,twocolumn,letterpaper]{article}
\usepackage[accsupp]{axessibility}
\usepackage{iccv}
\usepackage{times}
\usepackage{epsfig}
\usepackage{graphicx}
\usepackage{amsmath}
\usepackage{amssymb}
\usepackage{booktabs}
\usepackage{bbm}
\usepackage{bbding}
\usepackage{multirow}

\usepackage[pagebackref=true,breaklinks=true,letterpaper=true,colorlinks,bookmarks=false]{hyperref}

\iccvfinalcopy 

\def\iccvPaperID{5465} 

\ificcvfinal\fi

\begin{document}

\title{CLIP2Point: Transfer CLIP to Point Cloud Classification with Image-Depth Pre-Training}

\author{
Tianyu Huang$^{1,3}$ \quad Bowen Dong$^{1}$ \quad Yunhan Yang$^{1,4}$ \quad Xiaoshui Huang$^{2}$ \quad \\
Rynson W.H. Lau$^{3}$ \quad Wanli Ouyang$^{2}$ \quad Wangmeng Zuo$^{1,5}$\footnotemark[2] \\
$^{1}$Harbin Institute of Technology \quad $^{2}$Shanghai AI Laboratory \quad $^{3}$City University of Hong Kong \\ $^{4}$The University of Hong Kong \quad $^{5}$Peng Cheng Laboratory \\
{\tt\small tyhuang0428@gmail.com, rynson.lau@cityu.edu.hk, wanli.ouyang@sydney.edu.au, wmzuo@hit.edu.cn}
}

\maketitle
\ificcvfinal\thispagestyle{empty}\fi

\newcommand\blfootnote[1]{%
  \begingroup
  \renewcommand\thefootnote{}\footnote{#1}%
  \addtocounter{footnote}{-1}%
  \endgroup
}
\blfootnote{$^\dagger$Corresponding Author: Wangmeng Zuo (wmzuo@hit.edu.cn)}

\begin{abstract}
   Pre-training across 3D vision and language remains under development because of limited training data. Recent works attempt to transfer vision-language (V-L) pre-training methods to 3D vision. However, the domain gap between 3D and images is unsolved, so that V-L pre-trained models are restricted in 3D downstream tasks. To address this issue, we propose CLIP2Point, an image-depth pre-training method by contrastive learning to transfer CLIP to the 3D domain, and adapt it to point cloud classification. We introduce a new depth rendering setting that forms a better visual effect, and then render 52,460 pairs of images and depth maps from ShapeNet for pre-training. The pre-training scheme of CLIP2Point combines cross-modality learning to enforce the depth features for capturing expressive visual and textual features and intra-modality learning to enhance the invariance of depth aggregation. Additionally, we propose a novel Gated Dual-Path Adapter (GDPA), i.e., a dual-path structure with global-view aggregators and gated fusion for downstream representative learning. It allows the ensemble of CLIP and CLIP2Point, tuning pre-training knowledge to downstream tasks in an efficient adaptation. Experimental results show that CLIP2Point is effective in transferring CLIP knowledge to 3D vision. CLIP2Point outperforms other 3D transfer learning and pre-training networks, achieving state-of-the-art results on zero-shot, few-shot, and fully-supervised classification. Codes are available at: \href{https://github.com/tyhuang0428/CLIP2Point}{https://github.com/tyhuang0428/CLIP2Point}.
\end{abstract}

\section{Introduction}
Vision-language (V-L) pre-training has achieved great success in computer vision.
Benefiting from large-scale data, V-L pre-trained models~\cite{radford2021learning,yao2021filip} transfer language knowledge to visual understanding, which can be fine-tuned to multiple downstream tasks.
However, pre-training across 3D vision and language remains an open question, due to the lack of sufficient training data.
For example, Contrastive Language-Image Pre-training (CLIP)~\cite{radford2021learning} takes more than 400M image-text pairs as training data.
In contrast, few studies have been given to pre-training across 3D vision and language.
Moreover, even the conventional 3D pre-training method PointContrast~\cite{xie2020pointcontrast} is trained on ScanNet~\cite{dai2017scannet} with only 100k pairs of point clouds from 1,513 scenes. 
Due to the limitation of 3D pre-training, most existing 3D deep networks~\cite{qi2017pointnet++,wang2019dynamic} are trained from scratch on specific downstream datasets.

\begin{figure}[t]
    \centering
    \includegraphics[width=0.47\textwidth]{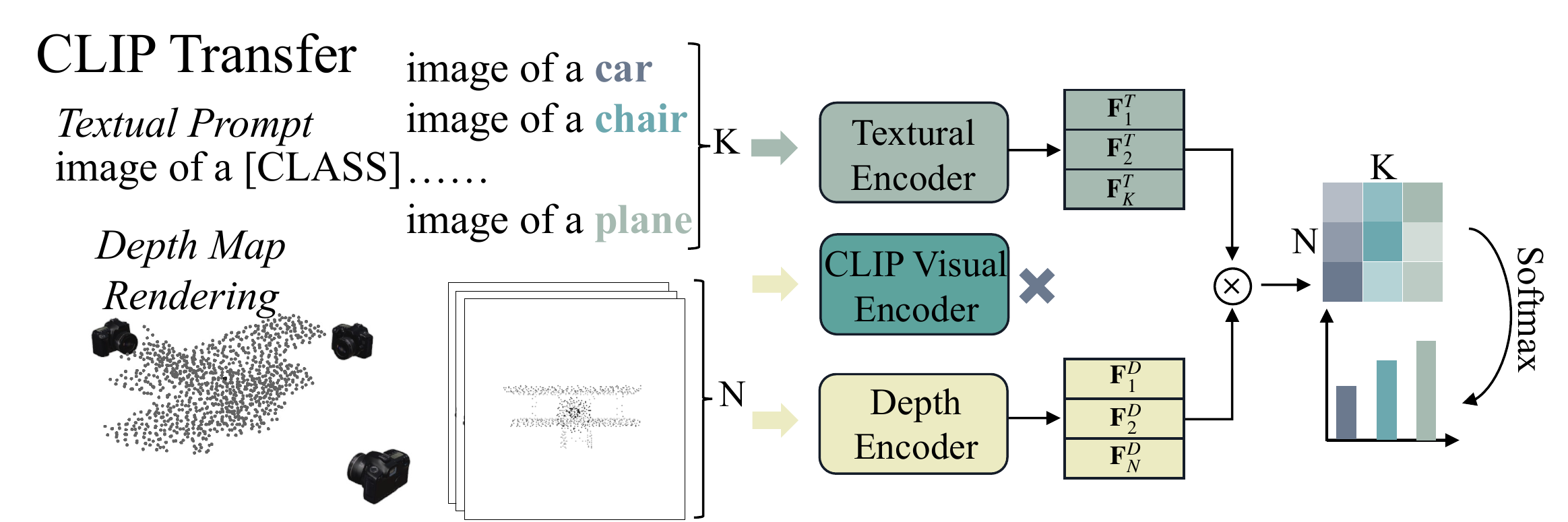}
    \caption{Overall architecture of CLIP transfer learning on the 3D domain. Point clouds are first projected to multi-view depth maps, and then aggregated by the CLIP visual encoder. Comparison with textual prompts presents the classification prediction. However, we argue that the domain gap exists between depth maps and CLIP pre-training images. To this end, a pre-trained depth encoder via CLIP2Point is proposed.}
\end{figure}

One remedy is to leverage the existing successful V-L pre-trained model for 3D vision tasks. To this end, one may first convert the 3D point clouds to multi-view 2D depth maps~\cite{su2015multi,goyal2021revisiting,hamdi2021mvtn,wang2022p2p}.
By simply treating 2D depth maps as images, PointCLIP~\cite{zhang2022pointclip} applies CLIP to 3D tasks, providing zero-shot and few-shot settings in the point cloud classification with textual prompting.
However, its results are still limited since the rendered depth maps are much different from the image domain of the CLIP training dataset.
And the sparsity and disorder of point cloud data result in various depth distributions from multiple views, further confusing the aggregation of CLIP.
Existing pre-training works focus on the domain gap~\cite{afham2022crosspoint} or multi-view consistency~\cite{xie2020pointcontrast} of point clouds, while we intend to tackle similar issues based on depth maps.
In addition, a solution of adapting pre-training knowledge to downstream tasks should be included in the V-L transfer.

In order to transfer CLIP to the 3D domain, we propose CLIP2Point, a pre-training scheme with two learning mechanisms: 1) \emph{cross-modality learning} for the contrastive alignment of RGB image and depth map, 2) \emph{intra-modality learning} in the depth modality to enhance the invariance of depth aggregation.
In particular, the image encoder $E_i$ is directly from CLIP weights and is frozen during pre-training. 
While the depth encoder $E_d$ is trained to 1) align depth features with CLIP image features in cross-modality learning and 2) encourage the depth aggregation to be invariant to view changes in intra-modality learning. With pre-training, the depth features can then be well aligned with the  visual CLIP features.
As for the training data, we do not adopt the depth maps in the existing RGB-D datasets as they are densely sampled and are contradicted to the sparsity of rendered depth maps. 
Instead, we reconstruct multi-view images and depth maps from 3D models directly.
Specifically, we render 10 views of RGB images from ShapeNet~\cite{chang2015shapenet}, which covers 52,460 3D models for 55 object categories. Meanwhile, we generate corresponding depth maps, with a new rendering setting that forms a better visual effect for CLIP encoding.
Experiments show that our CLIP2Point can significantly improve the performance of zero-shot point cloud classification. 

To further adapt our CLIP2Point to downstream tasks, we propose a novel Gated Dual-Path Adapter (GDPA).
Since our pre-training is to align the instance-level depth map, it can be complementary with CLIP pre-training knowledge that focuses on category-level discrimination.
We propose a dual-path structure, where both our pre-trained depth encoder $E_d$ and the CLIP visual encoder $E_i$ are utilized. A learnable global-view aggregator is attached to each encoder to extract an overall feature from multiple views. And the final logits can be calculated by the gated fusion of two encoders.

To sum up, our contributions can be summarized as:
\begin{itemize}
\item We propose a contrastive learning method dubbed CLIP2Point, with a newly proposed pre-training dataset that is pre-processed from ShapeNet, transferring CLIP knowledge to the 3D domain.
%
Experiments show that CLIP2Point significantly improves the performance of zero-shot classification.
\item We propose a novel Gated Dual-Path Adapter (GDPA), a dual-path structure with global-view aggregators and gated fusion to efficiently extend CLIP2Point to downstream representation learning.
\item Extensive experiments are conducted on ModelNet10, ModelNet40, and ScanobjectNN. %
In comparison to 3D transfer learning and pre-training networks, CLIP2Point achieves state-of-the-art results on zero-shot, few-shot, and fully-supervised point cloud classification tasks.
\end{itemize}
\section{Related Work}
\subsection{Vision-Language Pre-Training}
Vision-language (V-L) pre-training has been a growing interest in multi-modal tasks.
Pre-trained by large-scale image-text~\cite{chen2020uniter,chen2023vlp} or video-text~\cite{sun2019videobert} pairs, those models can be applied to multiple downstream tasks, \textit{e.g.}, visual question answering, image/video captioning, and text-to-image generation. CLIP~\cite{radford2021learning} further leverages V-L pre-training to transfer cross-modal knowledge, allowing natural language to understand visual concepts.
Nonetheless, pre-training across 3D vision and language~\cite{yao20223d,huang2022frozen} is restricted by insufficient 3D-text data pairs. And 3D downstream tasks like shape retrieval~\cite{han2019y2seq2seq} and text-guided shape generation~\cite{liu2022towards} suffer from limited performance.
Considering the vacancy between 3D vision and language, we attempt to transfer CLIP pre-trained knowledge to the 3D domain, making language applicable to point cloud classification.

\subsection{Self-Supervised Pre-Training}
Self-supervised pre-training has become an important issue in computer vision. Since task-related annotations are not required, it can leverage large-scale data and pretext tasks to learn general representation.
In particular, contrastive learning~\cite{he2020momentum,chen2020simple,mei2022unsupervised,wang2023mvcontrast} and masked auto-encoding~\cite{he2022masked,zhou2021ibot,devlin2018bert} are two popular self-supervised schemes.
Different from directly applying masked auto-encoding to 3D point completion~\cite{yu2022point,pang2022masked}, Li and Heizmann~\cite{li2022closer} argue that contrastive learning in 3D vision can vary from granularity (point/instance/scene) or modality (point/depth/image).
In this work, we aim to adopt image-depth contrastive learning to bridge the domain gap between depth features and visual CLIP features, thereby allowing to transfer CLIP knowledge to the 3D domain.

\begin{figure*}[t]
  \centering
  \includegraphics[width=0.85\textwidth]{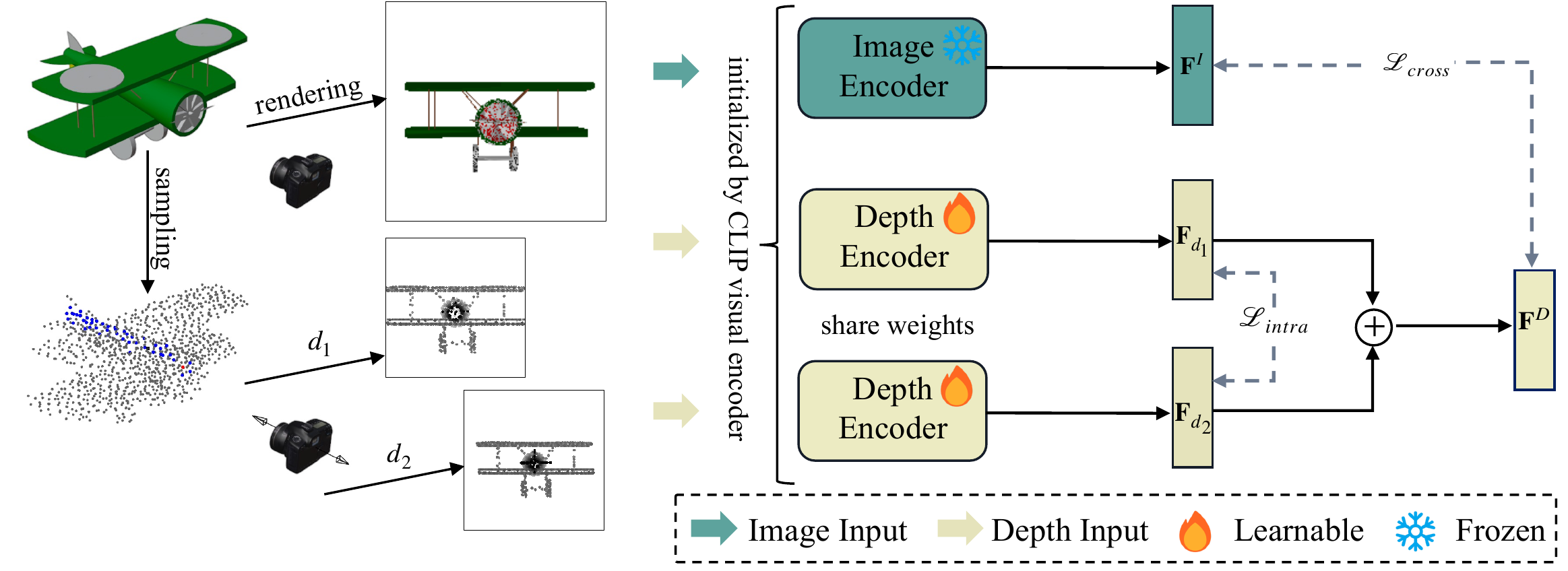}
  \caption{Pre-training scheme of CLIP2Point. 
  We propose a self-supervised pre-training scheme with intra-modality and cross-modality contrastive learning to align depth features with CLIP visual features. We randomly choose a camera view for each 3D model and modify the distances of the view to construct a pair of rendered depth maps. We adopt one NT-Xent loss between pairs of depth features extracted from the depth encoder and the other between image features and average depth features. We freeze the image encoder during training, enforcing the depth features by depth encoder to be aligned with the image features by CLIP visual encoder. Additionally, instead of all the blue points, we only consider the red point during depth rendering, which improves the visual effect.}
  \label{fig:pretrain}
\end{figure*}

\subsection{Downstream Fine-Tuning}
Fine-tuning has been widely used in downstream tasks to fit pre-trained weights to specific training datasets~\cite{zhai2019largescale,lin2014microsoft,zhou2017scene,zhang2022mining}. One common practice is to update the entire parameters during training, while it may be overfitted if the scale of training data is limited. Instead, partial tuning~\cite{cai2020tinytl,zhang2021tip} is a data-efficient way to fit downstream data. Recently, prompt tuning has been applied to language~\cite{brown2020language,li2021prefix} and vision~\cite{dosovitskiy2020image,jia2022visual} models. Prompt tuning provides several learnable token sequences and specific task heads for the adaptation, without the full tuning of pre-trained parameters.
Note that pre-trained models in 3D vision are still in early exploration, and existing deep networks in point cloud~\cite{qi2017pointnet++,wang2019dynamic,mohammadi2021pointview} all follow a full tuning paradigm.
In contrast, we propose a novel Gated Dual-Path Adapter for lightweight fine-tuning. With CLIP textual prompts, a supervised downstream setting is available by tuning efficient adapters only.
\section{CLIP-Based Transfer Learning in 3D}
Transfer learning works~\cite{su2015multi,goyal2021revisiting,hamdi2021mvtn,wang2022p2p} in 3D vision are basically based on 2D pre-training networks, converting point clouds to 2D depth maps. Recently, the success of V-L pre-training opens up potential opportunities for 3D-Language transfer. PointCLIP~\cite{zhang2022pointclip} directly adopts a CLIP visual encoder to projected depth maps. However, the image-depth domain gap restricts its performance. Instead, we manage to align depth features to the CLIP domain, allowing a boost on downstream tasks.

\subsection{Review of CLIP and PointCLIP}
\noindent\textbf{CLIP}~\cite{radford2021learning} is a vision-language pre-training method that matches images and texts by contrastive learning.
It contains two individual encoders: a visual encoder and a language encoder, to respectively extract image features $\mathbf{F}^I\in\mathbb{R}^{1 \times C}$ and textual features $\mathbf{F}^T\in\mathbb{R}^{1 \times C}$. Here, $C$ is the embedding dimension of encoders. For zero-shot transfer, the cosine similarity of $\mathbf{F}^I$ and $\mathbf{F}^T$ implies the matching results.
Taking a $K$-category classification task as an example, textual prompts are generated with the category names and then encoded by CLIP, extracting a list of textual features ${\{\mathbf{F}^{T}_k\}}_{k=1}^K\in\mathbb{R}^{K \times C}$. For each image feature $\mathbf{F}^I$, we can calculate the predicted probability $p$ as follows,
\begin{equation}
    \label{eq:logits}
    l_k =  \cos(\mathbf{F}^I, \mathbf{F}^{T}_k), \quad p = {\rm softmax}([l_1, \dots, l_K]).
\end{equation}
where $l_k$ denotes the logit for $k$-th category.

\noindent\textbf{PointCLIP}~\cite{zhang2022pointclip} applies CLIP to 3D point cloud data. It renders multi-view depth maps from point clouds, and then extracts the depth map features $\{\mathbf{F}_{v}^D\}_{v=1}^N$ with the CLIP visual encoder, where $N$ is the number of views. Logits of the zero-shot classification can be calculated similarly to Eq.(\ref{eq:logits}), while multi-view features are gathered with searched weights.
PointCLIP also proposes an inter-view adapter for the few-shot classification.
It adopts a residual form, which concatenates multi-view features $\{\mathbf{F}_{v}^D\}_{v=1}^N$ for a global representation $\mathbf{G}^{D}\in\mathbb{R}^{1 \times C}$ and then add $\mathbf{G}^{D}$ back to extract adapted features $\mathbf{\hat{F}}_{v}^D\in\mathbb{R}^{1 \times C}$. The adapter can be formulated as,
\begin{gather}
    \mathbf{G}^{D} = f_2({\rm ReLU}(f_1({\rm concat}(\{\mathbf{F}_{v}^D\}_{v=1}^N)))), \\
    \label{eq:global}
    \mathbf{\hat{F}}_{v}^D = {\rm ReLU}(\mathbf{G}^{D} \mathbf{W}_v^T), \\
    \label{eq:search}
    l_k = \sum_{v=1}^N \alpha_v (\cos((\mathbf{F}_{v}^D + \mathbf{\hat{F}}_{v}^D), \mathbf{F}^{T}_k)),
\end{gather}
where ${\rm concat}(\cdot)$ denotes the concatenation on channel dimensions, $f_1$ and $f_2$ are two-layer MLPs, and $\mathbf{W}_v\in \mathbb{R}^{C \times C}$ and $\alpha_v$ denote the view transformation and the summation weights of the $v$-th view. $f_1$, $f_2$, and $\mathbf{W}_v$ are learnable during the few-shot learning, while $\alpha_v$ is post-searched.

However, depth maps are representations of geometry information, which lack natural texture information.
Therefore, it is inappropriate to directly apply CLIP visual encoder for the extraction of depth features, leaving some leeway for boosting point cloud classification.
%

\subsection{Aligning with CLIP Visual Features}
Instead of directly applying CLIP visual encoder to depth maps, we suggest to learn a depth encoder for aligning depth features with CLIP visual features.  
%
%
In other words, we expect the extracted features of a rendered depth map to be consistent with CLIP visual features of the corresponding image.
Then, CLIP textual prompts can be directly adopted to match the depth features.
Moreover, since depth maps are presented in multiple views, the consistency of depth distribution needs maintaining as well.

Contrastive learning is a self-supervised pre-training method that aligns features of each sample with its positive samples, and satisfies our expectations of minimizing the distance between image and depth features, as well as enhancing the consistency of multi-view depth features.
We reconstruct a pre-training dataset from ShapeNet, which contains pairs of rendered RGB images and corresponding depth maps. We propose a pre-training scheme with intra-modality and cross-modality contrastive learning. Then, the pre-trained depth encoder can well adapt to CLIP prompts. To further generate depth maps with a better visual effect for CLIP encoding, a new depth rendering setting is adopted.

\subsubsection{Pre-Training Scheme}
As shown in Fig.~\ref{fig:pretrain}, our pre-training network includes a depth encoder $E_d$ and an image encoder $E_i$.
Given the input dataset $S = \{\mathbf{I}_i\}_{i=1}^{|S|}$, where $\mathbf{I}_i\in \mathbb{R}^{3 \times H \times W}$ is the $i$-th rendered image in a random camera view, we render the corresponding depth maps $\mathbf{D}_{i, d_1}$ and $\mathbf{D}_{i, d_2}$ in the same view angle with different distances $d_1$ and $d_2$.
We first adopt a intra-modality aggregation among $\{(\mathbf{D}_{i, d_1}, \mathbf{D}_{i, d_2})\}_{i=1}^{|S|}$ with $E_d$, and then extract image features from $\{\mathbf{I}_i\}_{i=1}^{|S|}$ with $E_i$, enforcing $E_d$ to keep consistent with $E_i$ in a cross-modality aspect.
$E_d$ and $E_i$ are both initialized with the weights of the visual encoder in CLIP.
We freeze the parameters of $E_i$ during training, while $E_d$ is learnable.

\noindent\textbf{Intra-Modality Learning.} Considering the sparsity and disorder of point clouds in the 3D space, even though we render depth maps at the same distance, distributions of depth values for different views vary a lot. To keep the invariance of distance aggregation in $E_d$, intra-modality contrastive learning is adopted.
For each input depth map $\mathbf{D}_i$, we randomly modify the distance of the camera view but keep the view angle, generating two augmented depth maps $\mathbf{D}_{i, d_1}$ and $\mathbf{D}_{i, d_2}$. $\mathbf{D}_{i, d_1}$ and $\mathbf{D}_{i, d_2}$ are then fed into $E_d$, extracting depth features $\mathbf{F}_{i, d_1}^D, \mathbf{F}_{i, d_2}^D \in \mathbb{R}^{1 \times C}$.
Following the NT-Xent loss in SimCLR~\cite{chen2020simple}, the intra-modality contrastive loss $\mathcal{L}_{intra}$ can be formulated as,
\begin{equation}
    \mathcal{L}_{intra} = \frac{1}{2N}\sum_{i=1}^{N} (l_{intra}^i(d_1, d_2) + l_{intra}^i(d_2, d_1)),
\end{equation}
where $N$ denotes the batch size. $l_{intra}^i(\cdot)$ is based on InfoNCE~\cite{oord2018representation} loss. Please refer to the supplementary for more details.
And the final depth feature map $\mathbf{F}^{D}_{i}$ is the mean of $\mathbf{F}_{i, d_1}^D$ and $\mathbf{F}_{i, d_2}^D$.

\noindent\textbf{Cross-Modality Learning.} For a set of rendered RGB-D data, cross-modality contrastive learning aims to minimize the distance between rendered images and depth maps in the same pair, while maximizing the distance of others. For each input image $\mathbf{I}_i$, we extract the image features $\mathbf{F}_{i}^I\in\mathbb{R}^{1 \times C}$, which is exactly the same as CLIP visual features.
Together with depth features $\mathbf{F}_{i}^D$, we obtain the cross-modality contrastive loss $\mathcal{L}_{cross}$ as follows,
\begin{equation}
    \label{eq:cross_loss}
    \mathcal{L}_{cross} = \frac{1}{2N}\sum_{i=1}^{N} (l_{cross}^i(D, I) + l_{cross}^i(I, D)).
\end{equation}
Similarly, $l_{cross}^i(\cdot)$ is based on InfoNCE~\cite{oord2018representation} loss.

$\mathcal{L}_{intra}$ and $\mathcal{L}_{cross}$ are independently propagated, and $\mathcal{L}_{intra}$ drops much faster than $\mathcal{L}_{cross}$ during our pre-training.
Thus, we adopt a multi-task loss~\cite{kendall2018multi} to balance the two terms. 
The overall loss function $\mathcal{L}$ is formulated as,
\begin{equation}
    \mathcal{L} = \frac{1}{\sigma^2} \mathcal{L}_{intra} + \mathcal{L}_{cross} + \log(\sigma + 1),
\end{equation}
where $\sigma$ is a learnable balance parameter.

\begin{figure*}[t]
  \centering
  \includegraphics[width=0.9\textwidth]{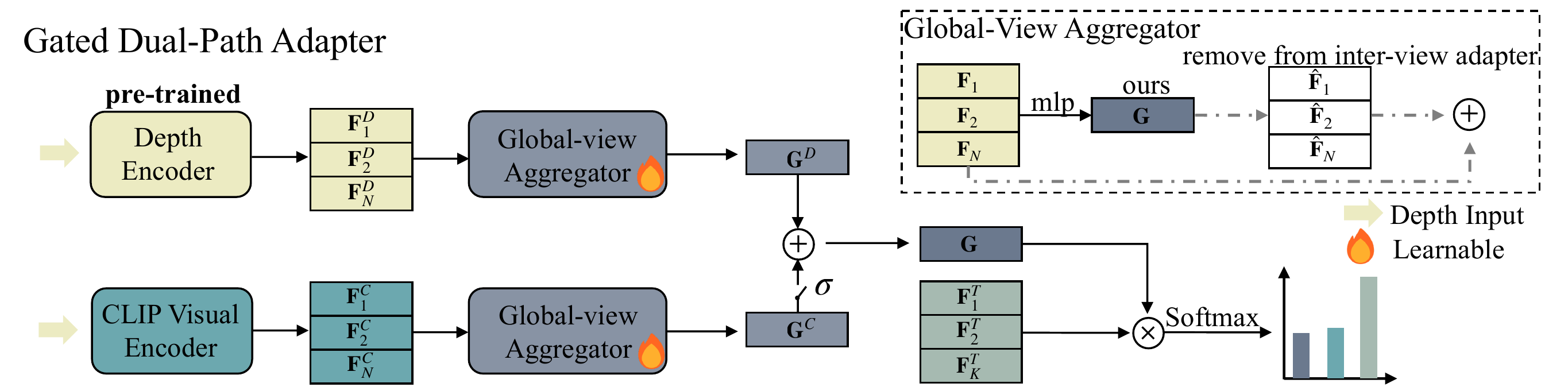}
  \caption{Gated Dual-Path Adapter (GDPA) for downstream learning. We design a dual-path structure, combining our pre-trained depth encoder with CLIP visual encoder. We propose a global-view aggregator and attach it to each encoder, which is parameter-efficient for downstream training. GDPA allows a fusion of knowledge in CLIP and our pre-training, enhancing the adaptation ability of CLIP2Point.}
  \label{fig:overview}
\end{figure*}

\subsubsection{Depth Rendering}
To convert point cloud data into rendered depth images,
we need to project 3D coordinates $(X, Y, Z)\in\mathbb{R}^3$ to 2D coordinates $(\hat{X}, \hat{Y})\in\mathbb{Z}^2$ in a specific view. 
%
Here we choose rendering from the front view as an example: a point at $(x, y, z)$ can simply match 
the corresponding pixel at $(\lceil x/z \rceil, \lceil y/z \rceil)$ by perspective projection. However, there are still two issues: 1) multiple points can be projected to the same pixel in a specific plane; 2) a large area of the rendered depth maps remains blank since points are sparsely distributed in the 3D space. For the first issue, existing works~\cite{goyal2021revisiting,zhang2022pointclip} prefer weighted summation of multiple points, 
\begin{equation}
    \label{eq:sum}
    d(\hat{x}, \hat{y}) = \frac{\sum_{(x, y, z)} z / (z + \epsilon)}{\sum_{(x, y, z)} 1 / z},
\end{equation}
where $(x, y, z)$ is the set of points matching $(\hat{x}, \hat{y})$, and $\epsilon$ denotes a minimal value, e.g., $1e^{-12}$.
We argue that the minimum depth value of those points is more intuitive in 2D vision, as we cannot watch an object perspectively with naked eyes.
For the second issue, few pixels can be covered due to the sparsity of point clouds. In order that the visual continuity of the depth value can be refined, we extend each point to its neighborhood pixels.
Taking $R$ as a dilation rate, we have the matching set $\mathbf{M}(\hat{x}, \hat{y}, R)$ as follows,
\begin{equation}
    \begin{aligned}
        \mathbf{M}(\hat{x}, \hat{y}, R) = \{(x, y, z)|(x, y, z) \in \mathbf{P} \quad and \\
        \hat{x} - \frac{R}{2} \leq \lceil \frac{x}{z} \rceil < \hat{x} + \frac{R}{2}, \hat{y} - \frac{R}{2} \leq \lceil \frac{y}{z} \rceil < \hat{y} + \frac{R}{2}\},
    \end{aligned}
\end{equation}
where $\mathbf{P}$ denotes the set of point clouds.
Previous rendering settings can be regarded as a special case when $R=1$. We set $R=2$, thus obtaining final rendered values as follows:
\begin{equation}
    \label{eq:render}
    d(\hat{x}, \hat{y}) = \min(z | (x, y, z) \in \mathbf{M}(\hat{x}, \hat{y}, 2)),
\end{equation}
where $\min(\cdot)$ denotes the minimum value of the input set.
We illustrate the rendering process in the bottom of Fig.~\ref{fig:pretrain}. We take the value of the red point in the airplane as the depth in $(0, 0)$, but previous works need to additionally consider all the blue points.

\subsection{Zero-Shot Classification}
With newly rendered depth maps and a pre-trained depth encoder, we can obtain better performance of zero-shot classification via a similar operation in CLIP.
And since depth features have a similar distribution to image features after pre-training, we can simply use the prompt, \textit{i.e.}, ``\textbf{image of a [class name]}" as the textual prompts.
After extracting depth features ${\{\mathbf{F}_{v}^D\}}_{v=1}^N$, we calculate the average logits of all the views as follows,
\begin{equation}
    \label{eq:avg}
    l_k = \frac{1}{N} \sum_{v=1}^N \cos(\mathbf{F}_{v}^D, \mathbf{F}^{T}_k).
\end{equation}
Note that PointCLIP exploits post-search to find a set of view weights $\{\alpha_v\}_{v=1}^N$ that achieves the highest accuracy. We argue that post-search is a time-consuming procedure, which is typically unfair for zero/few-shot tasks that require efficiency. Hence, we avoid post-search during training and evaluation, replacing it with the mean of multi-view logits.

\section{Downstream Representation Learning}
Albeit zero-shot learning is an efficient transfer pipeline to downstream tasks, lightweight fine-tuning is also useful for further refining the prediction accuracy.
To apply CLIP2Point to more tasks, we propose a novel Gated Dual-Path Adapter (GDPA) for representation learning. 

\subsection{Gated Dual-Path Adapter}
\noindent\textbf{Dual-Path Structure.}
CLIP2Point can achieve a significant improvement on zero-shot point cloud classification, as our pre-training narrows the domain gap between depth maps and images. While in few-shot learning, lightweight adapters also help transfer domains in a more direct way somehow, focusing on minimizing the category-level distance. That is the reason why PointCLIP can enjoy promising accuracy in few-shot classification. However, the domain transfer in our pre-training is based on instance-level discrimination, extracting and comparing global features. Thus, our pre-trained depth encoder and the CLIP visual encoder can be complementary, where the depth encoder can be adjusted to an appropriate feature domain, and the visual encoder can pay more attention to category selection.

\noindent\textbf{Global-View Aggregator.}
The ultimate goal of 3D-2D transferring is to aggregate a global representation of 3D objects from multi-view images.
While in PointCLIP, an inter-view adapter extracts global features and then expands them back to the dimension of input multi-view depth features $\mathbb{R}^{N \times C}$, which increases computational costs and the risk of information loss.
Moreover, remaining multi-view features still require aggregation, whether by time-consuming post-search or coarse feature averaging.
Instead, global features of multiple views can directly estimate a global logits vector. 
Therefore, we propose the global-view aggregator $g$:
\begin{equation}
g(\{\mathbf{F}_{v}\}_{v=1}^N) = f_2({\rm ReLU}(f_1({\rm concat}(\{\mathbf{F}_{v}\}_{v=1}^N)))),
\end{equation}
where $f_1$ and $f_2$ are linear layers. 
We can then reduce the learnable parameters and avoid post-search, presenting an efficient feature aggregation as follows:
\begin{equation}
\mathbf{G}^{C} = g(\{\mathbf{F}_{v}^{C}\}_{v=1}^N), \mathbf{G}^{D} = g(\{\mathbf{F}_{v}^{D}\}_{v=1}^N),
\end{equation}
where superscripts $C$ of features are related to CLIP. 

\noindent\textbf{Gated Fusion.}
Attention mechanisms are a common strategy in multi-modal interaction modules~\cite{tan2019lxmert}. However, a lightweight adapter cannot afford such a large computation.
A gate~\cite{ren2018gated} acts as a threshold to distinguish when to build an identity connection between layers, which can be naturally applied to the interaction of multiple layers.
To control the fusion of our multi-modal knowledge in an efficient way, we propose a gating strategy that adds learnable gating weights $\sigma$ to CLIP features. Then, we can calculate the final global feature $\mathbf{G}$ and $\rm logits$ as follows,
\begin{equation}
    \mathbf{G} = \sigma \cdot \mathbf{G}^{C} + \mathbf{G}^{D}, \quad l_k = \cos (\mathbf{G}, \mathbf{F}^{T}_k).
\end{equation}

\subsection{Downstream Supervision}
GDPA extends CLIP2Point to supervised downstream learning. Both few-shot and fully-supervised recognition tasks are available, and even scene-level tasks can be accessible if possible proposals are provided. Since GDPA is an object-level discrimination, we can simply use cross-entropy~\cite{de2005tutorial} loss for the supervision.
\section{Experiments}

\subsection{Datasets}
\noindent\textbf{Pre-Training Datasets.}
Numerous RGB-D datasets are available now, while depth images in those datasets cannot replace rendered depth maps, as they are densely annotated. To align images with sparsely marked depth maps, we have to directly convert 3D point clouds to depth maps.
ShapeNet~\cite{chang2015shapenet} is a large-scale dataset of 3D shapes, with 52,460 3D models in 55 categories. Previous works~\cite{xu2019disn,choy20163d} render a subset of ShapeNet in limited views. Instead, we render RGB images in 10 views with shapes and texture information from the complementary set of ShapeNet. The implementation follows MVTN~\cite{hamdi2021mvtn} on Pytorch3D~\cite{lassner2020pulsar}. Meanwhile, we sample the farthest 1,024 points of corresponding 3D models, and then render those points to depth maps as Eq.(\ref{eq:render}). To access the CLIP representation, the size of rendered images and depth maps is $224 \times 224$.
Following the separation of the classification benchmark on ShapeNet, we have 41,943 pairs for training and 10,517 pairs for validation.
For each training sample in the batch, we randomly choose a view out of the ten views. To evaluate the rendering quality, we conduct zero-shot classification experiments. The accuracy of RGB images and depth maps in our validation set are 54.21\% and 19.98\%, respectively.

\noindent\textbf{Downstream Datasets.}
We evaluate zero-shot classification on ModelNet10~\cite{wu20153d}, ModelNet40~\cite{wu20153d}, and ScanObjectNN~\cite{uy2019revisiting}, and 16-shot and fully-supervised classification on ModelNet40. ModelNet is a synthetic indoor 3D dataset, where ModelNet10 and ModelNet40 are both its subsets for classification. ModelNet10 contains 4,899 CAD models from 10 categories, including 3,991 for training and 908 for testing. While ModelNet40 contains 12,311 CAD models from 40 categories, with 9,843 for training and 2,468 for testing. Since the original ModelNet40 is not aligned in orientation, we use the aligned version~\cite{sedaghat2016orientation}. ScanObjectNN is a real-world dataset, which contains 2,902 samples of point cloud data from 15 categories. Different from clean CAD models in ModelNet, objects in ScanObjectNN are partially presented and attached with backgrounds. Thus, it is much harder than ModelNet. For all three datasets, we sample 1,024 points of each model as the input point cloud.

\begin{table}[t]
    \centering
    \caption{Quantitative results of zero-shot classification. Our pre-training significantly improves the accuracy, especially on ModelNet10, ModelNet40, and S-OBJ\_ONLY.}
    \begin{tabular}{l|ccc}
        \hline
        Models & PointCLIP\hspace{-3mm} & Ours w/o Pre.\hspace{-3mm} & Ours w/ Pre. \\
        \hline
        ModelNet10 & 30.23\hspace{-3mm} & 30.51\hspace{-3mm} & 66.63 \\
        ModelNet40 & 20.18\hspace{-3mm} & 29.71\hspace{-3mm} & 49.38 \\
        S-OBJ\_ONLY & 15.15\hspace{-3mm} & 28.40\hspace{-3mm} & 35.46 \\
        S-OBJ\_BG & 12.74\hspace{-3mm} & 23.92\hspace{-3mm} & 30.46 \\
        S-PB\_T50\_RS & 14.12\hspace{-3mm} & 18.18\hspace{-3mm} & 23.32 \\
        \hline
    \end{tabular}
    \label{tab:zero-shot}
\end{table}

\subsection{Implementation Details}
W use the basic version of Vision Transformer~\cite{dosovitskiy2020image} with a patch size of 32 (namely ViT-B/32) as our visual encoders for the encoding of both image and depth.
In pre-training, we use LAMB~\cite{you2019large} optimizer with a weight decay of $1 \times 10^{-4}$ and initialize the learning rate to $6 \times 10^{-3}$. Our pre-training takes 100 epochs with a batch size of 256. We choose the checkpoint with the highest accuracy in our evaluation set as the final weights for downstream tasks.
In few-shot and fully-supervised learning, we use AdamW~\cite{loshchilov2017decoupled} optimizer with a weight decay of $1 \times 10^{-4}$ and initialize the learning rate to $1 \times 10^{-3}$. The training batch size is 32.

\subsection{Zero-Shot Classification}
To the best of our knowledge, PointCLIP is the only attempt to conduct zero-shot classification on the whole 3D dataset. Previous works~\cite{cheraghian2019zero,cheraghian2022zero} divide 3D datasets into two parts: ``seen'' and ``unseen'' categories. Models are trained on the former and evaluated on the latter, which is easier than our zero-shot task. To evaluate the effectiveness of our depth rendering setting and pre-training transfer, we compare PointCLIP on ModelNet10, ModelNet40, and ScanobjectNN. For ScanobjectNN, we test on the object-only split (S-OBJ\_ONLY), the background split (S-OBJ\_BG), and the hardest split (S-PB\_T50\_RS), respectively.

As shown in Tab.~\ref{tab:zero-shot}, even without pre-training, our method can outperform PointCLIP, simply by using the newly rendered depth maps. Especially on S-OBJ\_ONLY, we have an over 10\% gain on the accuracy, which means our rendered depth maps can draw close to the CLIP domain, even in real-world noisy data.
After pre-training, the accuracy is significantly improved on ModelNet10 and ModelNet40, by 36.40\% and 29.20\%. Nonetheless, a 20.31\% gain can also be attained on S-OBJ\_ONLY.
While the improvement on S-OBJ\_BG and S-PB\_T50\_RS is relatively small. We think that is because we generate our pre-training dataset from ShapeNet, which is a clean synthetic dataset without background. Background points may somewhat disturb object discrimination.
We also notice that PointCLIP has better accuracy on S-PB\_T50\_RS than on S-OBJ\_BG. S-PB\_T50\_RS is an augmented version, which contains much more cases than S-OBJ\_BG. Since both of the results are relatively low, we judge that its performance on real-world datasets like ScanObjectNN may be unstable.

\begin{table}[t]
    \centering
    \caption{Quantitative results of few-shot classification. Our few-shot pipeline has already achieved state-of-the-art results, and the pre-trained version can further improve the performance.}
    \begin{tabular}{l|c|cc}
        \hline
        Method & Encoder & w/o Pre. & w/ Pre. \\
        \hline
        CrossPoint & DGCNN & 81.56 & 84.48 \\
        Point-MAE & Transformer & 79.70 & 84.20 \\
        PointCLIP & ViT-B/32 & 83.83 & - \\
        PointCLIP & ResNet101 & 87.20 & - \\
        CLIP2Point & ViT-B/32 & 87.46 & 89.79 \\
        \hline
    \end{tabular}
    \label{tab:few-shot}
\end{table}

\subsection{16-Shot Classification}
To further evaluate the transfer ability of our pre-training and verify the Gated Dual-Path Adapter, we compare with PointCLIP, as well as two self-supervised pre-training methods: CrossPoint~\cite{afham2022crosspoint} and Point-MAE~\cite{pang2022masked} in 16-shot classification. We choose a DGCNN~\cite{wang2019dynamic} backbone for CrossPoint and a 12-layer Transformer encoder for Point-MAE. Although we only use ViT-B/32 as our encoder, PointCLIP in ResNet101 is included in the experiments.

We present the quantitative results of our few-shot experiments in Tab.~\ref{tab:few-shot}.
Initialized by CLIP weights, our few-shot pipeline w/o pre-training has already outperformed other methods, thanks to the global-view aggregator.
Moreover, our pre-trained version can reach an accuracy of 89.79\%, which is very close to some traditional supervised networks such as PointNet++~\cite{qi2017pointnet++}.


\subsection{Fully-Supervised Classification}
Although GDPA is a lightweight adapter that is fit for few-shot tasks, we can also apply it to fully-supervising.
We conduct fully-supervised classification experiments on ModelNet40, comparing with 5 state-of-the-art 3D transfer networks MVCNN~\cite{su2015multi}, SimpleView~\cite{goyal2021revisiting}, MVTN~\cite{hamdi2021mvtn}, PointCLIP~\cite{zhang2022pointclip}, and P2P~\cite{wang2022p2p}. Similar to CLIP2Point, these networks convert point cloud data to depth maps, leveraging 2D pre-trained backbones to extract corresponding shape features. We also compare with 3D self-supervised pre-training methods Point-BERT~\cite{yu2022point} and Point-MAE~\cite{pang2022masked}.

As shown in Tab.~\ref{tab:supervised}, CLIP2Point outperforms P2P (HorNet-L), but with much lower input requirements.  Additionally, our training only fine-tunes learnable adapters, which is more efficient than those full-tuning methods. Experimental results prove that we achieve state-of-the-art with low requirements of inputs and parameters.

\begin{figure*}
    \centering
    \begin{tabular}{rccccccc}
        airplane &
        \includegraphics[width=0.07\textwidth]{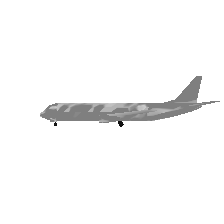} &
        \includegraphics[width=0.07\textwidth]{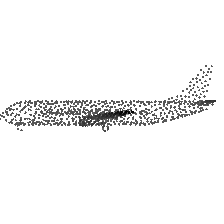} &
        \includegraphics[width=0.07\textwidth]{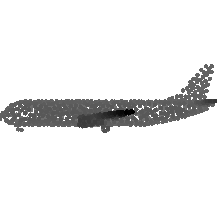} &
        \includegraphics[width=0.07\textwidth]{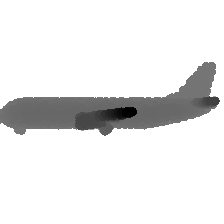} &
        \includegraphics[width=0.07\textwidth]{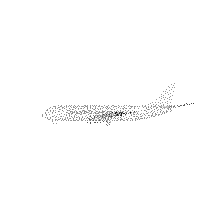} &
        \includegraphics[width=0.07\textwidth]{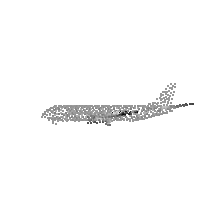} &
        \includegraphics[width=0.07\textwidth]{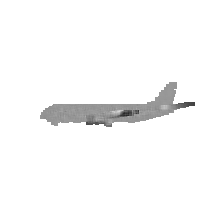} \\
        bench &
        \includegraphics[width=0.07\textwidth]{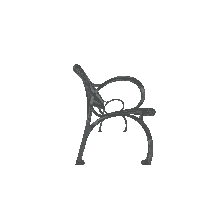} &
        \includegraphics[width=0.07\textwidth]{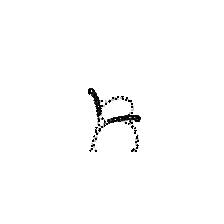} &
        \includegraphics[width=0.07\textwidth]{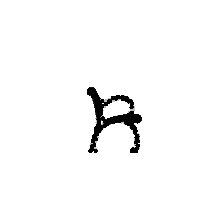} &
        \includegraphics[width=0.07\textwidth]{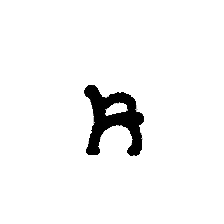} &
        \includegraphics[width=0.07\textwidth]{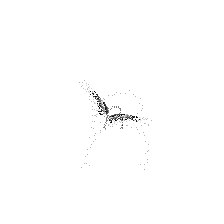} &
        \includegraphics[width=0.07\textwidth]{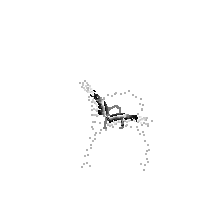} &
        \includegraphics[width=0.07\textwidth]{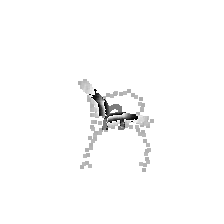} \\
        car &
        \includegraphics[width=0.07\textwidth]{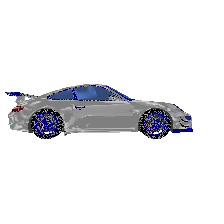} &
        \includegraphics[width=0.07\textwidth]{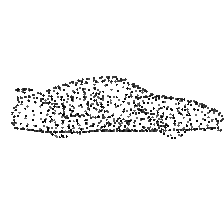} &
        \includegraphics[width=0.07\textwidth]{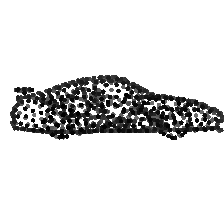} &
        \includegraphics[width=0.07\textwidth]{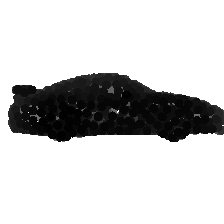} &
        \includegraphics[width=0.07\textwidth]{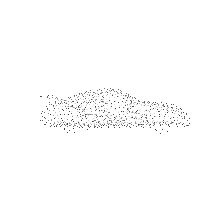} &
        \includegraphics[width=0.07\textwidth]{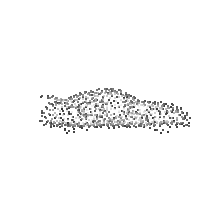} &
        \includegraphics[width=0.07\textwidth]{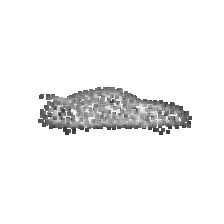} \\
        & rendered & Minimum & Minimum & Minimum & Weighted & Weighted & Weighted\\
        & image & $R=1$ & $R=2$ & $R=4$ & $R=1$ & $R=2$ & $R=4$\\
    \end{tabular}
    \caption{Visualization results of our rendered images with different rendering settings.}
    \label{fig:vis3}
\end{figure*}


\begin{table}[t]
    \centering
    \caption{Fully-supervised classification on ModelNet40. Data Type presents numbers of input points and views respectively.}
    \begin{tabular}{l|c|c}
        \hline
        Methods & Data Type & Acc.(\%) \\
        \hline
        MVCNN & image, 12 & 90.1 \\
        SimpleView & 1,024, 6 & 93.4 \\
        MVTN & 2,048, 12 & 93.8 \\
        PointCLIP & 1,024, 10 & 92.1 \\
        P2P: ResNet-101 & 4,096, 40 & 93.1 \\
        P2P: ConvNeXt-L & 4,096, 40 & 93.2 \\
        P2P: HorNet-L & 4,096, 40 & 94.0 \\
        \hline
        Transformer & 1,024, - & 91.4 \\
        + Point-BERT & 1,024, - & 93.2 \\
        + Point-MAE & 1,024, - & 93.8 \\
        \hline
        CLIP2Point (Ours) & 1,024, 10 & \textbf{94.2} \\
        \hline
    \end{tabular}
    \label{tab:supervised}
\end{table}

\subsection{Ablation Study}
\noindent\textbf{Intra-Modality Learning.}
In CLIP2Point, cross-modality learning is necessary to bridge the image-depth domain gap, while intra-modality learning is an extra enhancement of depth invariance.
To evaluate the effectiveness of our intra-modality learning, we conduct a pre-training experiment with cross-modality only, in which the accuracy of zero-shot classification is only \textbf{38.29\%}.
Regardless of random view distances, we simply extract the features of original depth maps as $\mathbf{F}_{i}^D$. The final loss can be formulated as Eq.(\ref{eq:cross_loss}). We keep the same pre-training setting, while the result of zero-shot classification in this version of pre-training is \textbf{11.09\% lower} than the version with intra-modality.
Our intra-modality contrastive learning allows the depth encoder to keep a depth invariance among different camera views. Without randomized distances and corresponding contrastive restrictions, the encoder may easily fail when depth values vary a lot in different views.


\noindent\textbf{Gated Dual-Path Adapter.}
To evaluate the design of the Gated Dual-Path Adapter, we experiment with the dual-path structure, global-view aggregators, and gated weights. For the single-path adapter, we extract features with a pre-trained depth encoder in CLIP2Point. 

As shown in Tab.~\ref{tab:adapter}, all the components play a role in the improvement. Extra weights in the inter-view adapter make few-shot training much easier to overfit. And both the expanding of global features and the gathering of multi-view features may cause information loss. Additionally, the dual-path structure greatly improves the performance in our global-view aggregator, which demonstrates that knowledge from CLIP2Point and CLIP are complementary. While the improvement in the inter-view adapter is relatively small, we think multi-view gathering in a single encoder may disturb the combination of encoders.

We further evaluate a GDPA, in which encoders of both paths are initialized with the CLIP visual encoder. Its accuracy is \textbf{87.20\%}, which is even lower than the single-path version (87.46\%, Tab.~\ref{tab:few-shot}). It means the improvement of the dual-path structures comes from the fine fusion of pre-training knowledge, rather than increased parameters.

\noindent\textbf{Depth Rendering.}
To analyze the depth rendering setting, we evaluate several settings in Tab.~\ref{tab:depth}. ``Weighted'' and ``Minimum'' represent the depth values described in Eq.(\ref{eq:sum}) and Eq.(\ref{eq:render}), respectively. The dilation rate is set to 1, 2, and 4 for ablation studies. To fairly compare all the settings, we use a single-path adapter without pre-training.

As shown in Tab.~\ref{tab:depth}, using the minimum depth value has much higher accuracy in both zero-shot and few-shot classification. We think that is because the visual effects in the ``Minimum'' depth value can be close to CLIP pre-training images. However, the larger is not the better for the dilation rate. A too-large dilation rate may blur depth maps, especially near the corners of objects. The visualization in Fig.~\ref{fig:vis3} further demonstrates that our setting has the best visual effect. It seems a global-view aggregator with our depth rendering is the main growth point in few-shot. While we argue that, without a pre-training ensemble, it can hardly achieve accuracy close to fully-supervised networks, since we only use 16 samples per category for few-shot learning.

\begin{figure}[!htbp]
    \centering
    \includegraphics[width=0.45\textwidth]{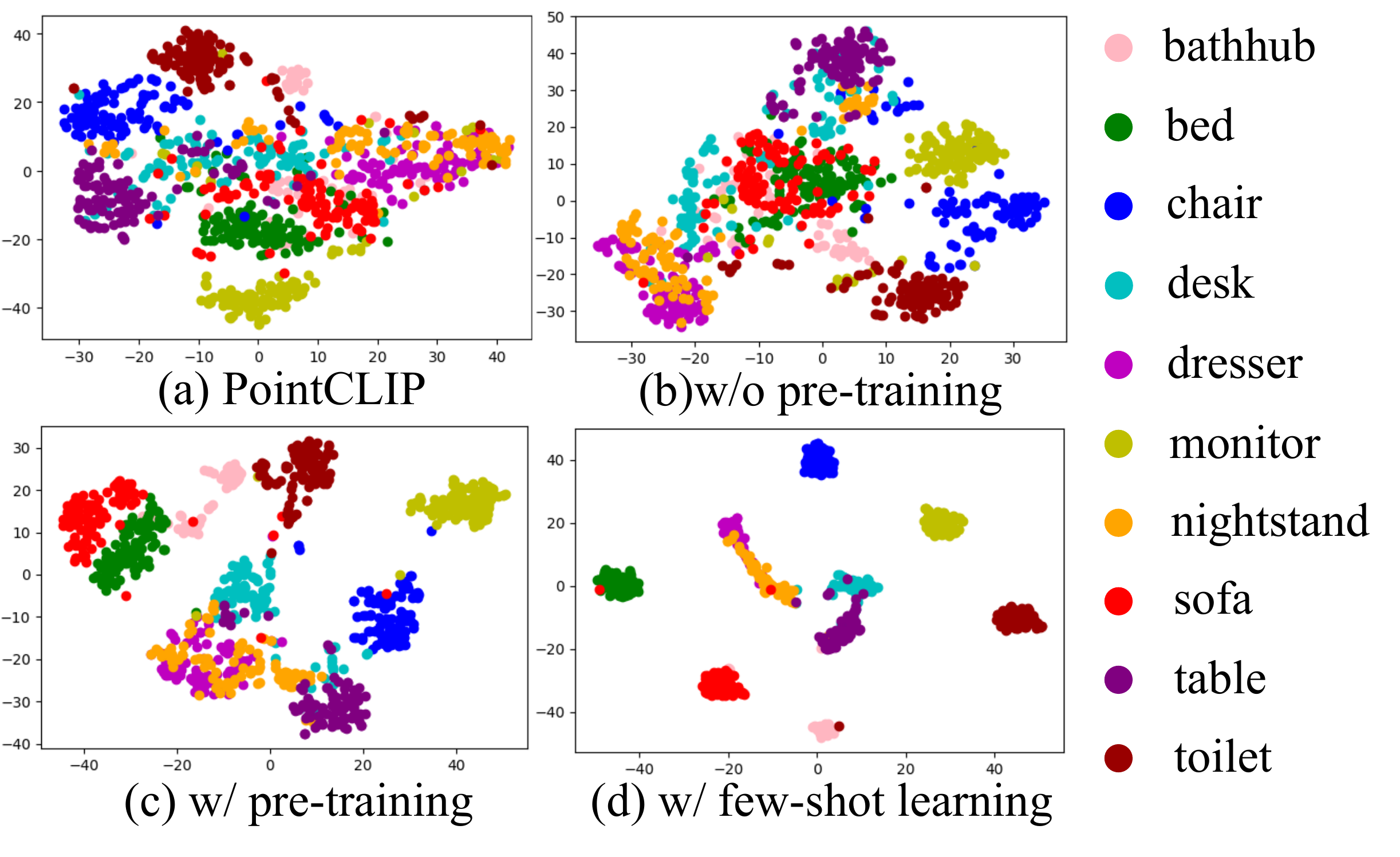}
    \caption{Visualization of feature distributions on ModelNet10.}
    \label{fig:vis_feat}
\end{figure}

\noindent\textbf{Visualization of Feature Distribution.}
To further verify the effectiveness of CLIP2Point, we visualize features on ModelNet10 in Fig.~\ref{fig:vis_feat}. Features extracted by PointCLIP (a) and CLIP2Point w/o pre-training (b) are both chaotic. After pre-trained, CLIP2Point (c) works well in separating classes in feature space, which indicates that our pre-training greatly improves the feature representation. However, there still exist two similar categories that are mixed up. After few-shot training, all classes are clearly separated. It is consistent with the results in the main text.

\begin{table}[!htbp]
    
    \centering
    \caption{Few-shot classification with different components.}
    \label{tab:adapter}
    \begin{tabular}{c|cc}
        \hline
        \scriptsize
        Dual-Path & \scriptsize Inter-view & \scriptsize Global-view \\
        \hline
        \scriptsize
        \XSolidBrush & \scriptsize 86.06 & \scriptsize 87.32 \\
        \scriptsize
        \checkmark & \scriptsize 86.18 & \scriptsize 89.21 (w/o gating) / 89.79 (gating) \\
        \hline
    \end{tabular}
    \vspace{-1.3em}
\end{table}

\begin{table}[!htbp]
    \centering

    \caption{Classification performance in different rendering settings.}
   
    \label{tab:depth}
    
    \begin{tabular}{l|c|ccc}
        \hline
        \scriptsize Task & \scriptsize Depth Value & \scriptsize R=1 & \scriptsize R=2 & \scriptsize R=4 \\
        \hline
        \multicolumn{1}{l|}{\multirow{2}{*}{\scriptsize Zero-Shot}} & \scriptsize Weighted & \scriptsize 16.86 & \scriptsize 17.63 & \scriptsize 21.11 \\
        \multicolumn{1}{l|}{} & \scriptsize Minimum & \scriptsize 24.87 & \scriptsize 29.71 & \scriptsize 28.36 \\
        \hline
        \multicolumn{1}{l|}{\multirow{2}{*}{\scriptsize Few-Shot}} & \scriptsize Weighted & \scriptsize 83.91 & \scriptsize 83.87 & \scriptsize 84.08 \\
        \multicolumn{1}{l|}{} & \scriptsize Minimum & \scriptsize 86.47 & \scriptsize 87.46 & \scriptsize 87.40 \\
        \hline
    \end{tabular}
    \vspace{-1.2em}
\end{table}
\section{Conclusion}
In this paper, we propose CLIP2Point, which pre-trains a depth encoder for adapting CLIP knowledge to the 3D domain. We introduce a depth-image pre-training method, which consists of both intra-modality and cross-modality contrastive learning to bridge the domain gap between depth features by depth encoder and image features by  CLIP visual encoder, and to maintain the invariance of multi-view depth distribution.
For the pre-training data, we render 52,560 images from 3D models in ShapeNet, and meanwhile generate corresponding depth maps with a new depth rendering setting. After pre-training, the performance of zero-shot point cloud classification is significantly improved. To further adapt our pre-trained weights to downstream tasks, we newly propose Gated Dual-Path Adapter. With global-view aggregators and gated fusion in a dual-path structure, we achieve state-of-the-art results in comparison with 3D transfer learning and pre-training methods.

CLIP2Point is effective in transferring CLIP knowledge to 3D vision,  opening up the possibilities to apply CLIP for zero-shot and open-world 3D tasks. Moreover, a similar idea can also be used to transfer CLIP knowledge to other visual modalities such as near-infrared (NIR) images. Nonetheless, its performance and generalization ability are affected by the amount and quality of pre-training data. For example, synthetic pre-training data is limited to real-world downstream tasks that contain noise and complicated background information. In future, we will leverage more realistic data in pre-training, and extend CLIP2Point to other complex 3D tasks (e.g., detection and segmentation).

\section*{Acknowledgements}
This work was supported by National Key RD Program of China under Grant No. 2021ZD0112100, and the National Natural Science Foundation of China (NSFC) under Grant No. U19A2073.

{\small
\bibliographystyle{ieee_fullname}
\bibliography{egbib}

\begin{thebibliography}{10}\itemsep=-1pt

\bibitem{afham2022crosspoint}
Mohamed Afham, Isuru Dissanayake, Dinithi Dissanayake, Amaya Dharmasiri,
  Kanchana Thilakarathna, and Ranga Rodrigo.
\newblock Crosspoint: Self-supervised cross-modal contrastive learning for 3d
  point cloud understanding.
\newblock In {\em Proceedings of the IEEE/CVF Conference on Computer Vision and
  Pattern Recognition}, pages 9902--9912, 2022.

\bibitem{brown2020language}
Tom Brown, Benjamin Mann, Nick Ryder, Melanie Subbiah, Jared~D Kaplan, Prafulla
  Dhariwal, Arvind Neelakantan, Pranav Shyam, Girish Sastry, Amanda Askell,
  et~al.
\newblock Language models are few-shot learners.
\newblock {\em Advances in neural information processing systems},
  33:1877--1901, 2020.

\bibitem{cai2020tinytl}
Han Cai, Chuang Gan, Ligeng Zhu, and Song Han.
\newblock Tinytl: Reduce memory, not parameters for efficient on-device
  learning.
\newblock {\em Advances in Neural Information Processing Systems},
  33:11285--11297, 2020.

\bibitem{chang2015shapenet}
Angel~X Chang, Thomas Funkhouser, Leonidas Guibas, Pat Hanrahan, Qixing Huang,
  Zimo Li, Silvio Savarese, Manolis Savva, Shuran Song, Hao Su, et~al.
\newblock Shapenet: An information-rich 3d model repository.
\newblock {\em arXiv preprint arXiv:1512.03012}, 2015.

\bibitem{chen2023vlp}
Fei-Long Chen, Du-Zhen Zhang, Ming-Lun Han, Xiu-Yi Chen, Jing Shi, Shuang Xu,
  and Bo Xu.
\newblock Vlp: A survey on vision-language pre-training.
\newblock {\em Machine Intelligence Research}, 20(1):38--56, 2023.

\bibitem{chen2020simple}
Ting Chen, Simon Kornblith, Mohammad Norouzi, and Geoffrey Hinton.
\newblock A simple framework for contrastive learning of visual
  representations.
\newblock In {\em International conference on machine learning}, pages
  1597--1607. PMLR, 2020.

\bibitem{chen2020uniter}
Yen-Chun Chen, Linjie Li, Licheng Yu, Ahmed El~Kholy, Faisal Ahmed, Zhe Gan, Yu
  Cheng, and Jingjing Liu.
\newblock Uniter: Universal image-text representation learning.
\newblock In {\em European conference on computer vision}, pages 104--120.
  Springer, 2020.

\bibitem{cheraghian2022zero}
Ali Cheraghian, Shafin Rahman, Townim~F Chowdhury, Dylan Campbell, and Lars
  Petersson.
\newblock Zero-shot learning on 3d point cloud objects and beyond.
\newblock {\em International Journal of Computer Vision}, pages 1--21, 2022.

\bibitem{cheraghian2019zero}
Ali Cheraghian, Shafin Rahman, and Lars Petersson.
\newblock Zero-shot learning of 3d point cloud objects.
\newblock In {\em 2019 16th International Conference on Machine Vision
  Applications (MVA)}, pages 1--6. IEEE, 2019.

\bibitem{choy20163d}
Christopher~B Choy, Danfei Xu, JunYoung Gwak, Kevin Chen, and Silvio Savarese.
\newblock 3d-r2n2: A unified approach for single and multi-view 3d object
  reconstruction.
\newblock In {\em Proceedings of the European Conference on Computer Vision
  ({ECCV})}, 2016.

\bibitem{dai2017scannet}
Angela Dai, Angel~X. Chang, Manolis Savva, Maciej Halber, Thomas Funkhouser,
  and Matthias Nie{\ss}ner.
\newblock Scannet: Richly-annotated 3d reconstructions of indoor scenes.
\newblock In {\em Proc. Computer Vision and Pattern Recognition (CVPR), IEEE},
  2017.

\bibitem{de2005tutorial}
Pieter-Tjerk De~Boer, Dirk~P Kroese, Shie Mannor, and Reuven~Y Rubinstein.
\newblock A tutorial on the cross-entropy method.
\newblock {\em Annals of operations research}, 134(1):19--67, 2005.

\bibitem{devlin2018bert}
Jacob Devlin, Ming-Wei Chang, Kenton Lee, and Kristina Toutanova.
\newblock Bert: Pre-training of deep bidirectional transformers for language
  understanding.
\newblock {\em arXiv preprint arXiv:1810.04805}, 2018.

\bibitem{dosovitskiy2020image}
Alexey Dosovitskiy, Lucas Beyer, Alexander Kolesnikov, Dirk Weissenborn,
  Xiaohua Zhai, Thomas Unterthiner, Mostafa Dehghani, Matthias Minderer, Georg
  Heigold, Sylvain Gelly, et~al.
\newblock An image is worth 16x16 words: Transformers for image recognition at
  scale.
\newblock {\em arXiv preprint arXiv:2010.11929}, 2020.

\bibitem{goyal2021revisiting}
Ankit Goyal, Hei Law, Bowei Liu, Alejandro Newell, and Jia Deng.
\newblock Revisiting point cloud shape classification with a simple and
  effective baseline.
\newblock {\em International Conference on Machine Learning}, 2021.

\bibitem{hamdi2021mvtn}
Abdullah Hamdi, Silvio Giancola, and Bernard Ghanem.
\newblock Mvtn: Multi-view transformation network for 3d shape recognition.
\newblock In {\em Proceedings of the IEEE/CVF International Conference on
  Computer Vision}, pages 1--11, 2021.

\bibitem{han2019y2seq2seq}
Zhizhong Han, Mingyang Shang, Xiyang Wang, Yu-Shen Liu, and Matthias Zwicker.
\newblock Y2seq2seq: Cross-modal representation learning for 3d shape and text
  by joint reconstruction and prediction of view and word sequences.
\newblock In {\em Proceedings of the AAAI Conference on Artificial
  Intelligence}, volume~33, pages 126--133, 2019.

\bibitem{he2022masked}
Kaiming He, Xinlei Chen, Saining Xie, Yanghao Li, Piotr Doll{\'a}r, and Ross
  Girshick.
\newblock Masked autoencoders are scalable vision learners.
\newblock In {\em Proceedings of the IEEE/CVF Conference on Computer Vision and
  Pattern Recognition}, pages 16000--16009, 2022.

\bibitem{he2020momentum}
Kaiming He, Haoqi Fan, Yuxin Wu, Saining Xie, and Ross Girshick.
\newblock Momentum contrast for unsupervised visual representation learning.
\newblock In {\em Proceedings of the IEEE/CVF conference on computer vision and
  pattern recognition}, pages 9729--9738, 2020.

\bibitem{huang2022frozen}
Xiaoshui Huang, Sheng Li, Wentao Qu, Tong He, Yifan Zuo, and Wanli Ouyang.
\newblock Frozen clip model is efficient point cloud backbone.
\newblock {\em arXiv preprint arXiv:2212.04098}, 2022.

\bibitem{jia2022visual}
Menglin Jia, Luming Tang, Bor-Chun Chen, Claire Cardie, Serge Belongie, Bharath
  Hariharan, and Ser-Nam Lim.
\newblock Visual prompt tuning.
\newblock {\em arXiv preprint arXiv:2203.12119}, 2022.

\bibitem{kendall2018multi}
Alex Kendall, Yarin Gal, and Roberto Cipolla.
\newblock Multi-task learning using uncertainty to weigh losses for scene
  geometry and semantics.
\newblock In {\em Proceedings of the IEEE conference on computer vision and
  pattern recognition}, pages 7482--7491, 2018.

\bibitem{lassner2020pulsar}
Christoph Lassner and Michael Zollh\"ofer.
\newblock Pulsar: Efficient sphere-based neural rendering.
\newblock {\em arXiv:2004.07484}, 2020.

\bibitem{li2022closer}
Lanxiao Li and Michael Heizmann.
\newblock A closer look at invariances in self-supervised pre-training for 3d
  vision.
\newblock {\em arXiv preprint arXiv:2207.04997}, 2022.

\bibitem{li2021prefix}
Xiang~Lisa Li and Percy Liang.
\newblock Prefix-tuning: Optimizing continuous prompts for generation.
\newblock {\em arXiv preprint arXiv:2101.00190}, 2021.

\bibitem{lin2014microsoft}
Tsung-Yi Lin, Michael Maire, Serge Belongie, James Hays, Pietro Perona, Deva
  Ramanan, Piotr Doll{\'a}r, and C~Lawrence Zitnick.
\newblock Microsoft coco: Common objects in context.
\newblock In {\em European conference on computer vision}, pages 740--755.
  Springer, 2014.

\bibitem{liu2022towards}
Zhengzhe Liu, Yi Wang, Xiaojuan Qi, and Chi-Wing Fu.
\newblock Towards implicit text-guided 3d shape generation.
\newblock In {\em Proceedings of the IEEE/CVF Conference on Computer Vision and
  Pattern Recognition}, pages 17896--17906, 2022.

\bibitem{loshchilov2017decoupled}
Ilya Loshchilov and Frank Hutter.
\newblock Decoupled weight decay regularization.
\newblock {\em arXiv preprint arXiv:1711.05101}, 2017.

\bibitem{lu2022open}
Yuheng Lu, Chenfeng Xu, Xiaobao Wei, Xiaodong Xie, Masayoshi Tomizuka, Kurt
  Keutzer, and Shanghang Zhang.
\newblock Open-vocabulary 3d detection via image-level class and debiased
  cross-modal contrastive learning.
\newblock {\em arXiv preprint arXiv:2207.01987}, 2022.

\bibitem{mei2022unsupervised}
Guofeng Mei, Xiaoshui Huang, Juan Liu, Jian Zhang, and Qiang Wu.
\newblock Unsupervised point cloud pre-training via contrasting and clustering.
\newblock In {\em 2022 IEEE International Conference on Image Processing
  (ICIP)}, pages 66--70. IEEE, 2022.

\bibitem{mohammadi2021pointview}
Seyed~Saber Mohammadi, Yiming Wang, and Alessio Del~Bue.
\newblock Pointview-gcn: 3d shape classification with multi-view point clouds.
\newblock In {\em 2021 IEEE International Conference on Image Processing
  (ICIP)}, pages 3103--3107. IEEE, 2021.

\bibitem{oord2018representation}
Aaron van~den Oord, Yazhe Li, and Oriol Vinyals.
\newblock Representation learning with contrastive predictive coding.
\newblock {\em arXiv preprint arXiv:1807.03748}, 2018.

\bibitem{pang2022masked}
Yatian Pang, Wenxiao Wang, Francis~EH Tay, Wei Liu, Yonghong Tian, and Li Yuan.
\newblock Masked autoencoders for point cloud self-supervised learning.
\newblock {\em arXiv preprint arXiv:2203.06604}, 2022.

\bibitem{qi2017pointnet++}
Charles~Ruizhongtai Qi, Li Yi, Hao Su, and Leonidas~J Guibas.
\newblock Pointnet++: Deep hierarchical feature learning on point sets in a
  metric space.
\newblock {\em Advances in neural information processing systems}, 30, 2017.

\bibitem{radford2021learning}
Alec Radford, Jong~Wook Kim, Chris Hallacy, Aditya Ramesh, Gabriel Goh,
  Sandhini Agarwal, Girish Sastry, Amanda Askell, Pamela Mishkin, Jack Clark,
  et~al.
\newblock Learning transferable visual models from natural language
  supervision.
\newblock In {\em International Conference on Machine Learning}, pages
  8748--8763. PMLR, 2021.

\bibitem{ren2018gated}
Wenqi Ren, Lin Ma, Jiawei Zhang, Jinshan Pan, Xiaochun Cao, Wei Liu, and
  Ming-Hsuan Yang.
\newblock Gated fusion network for single image dehazing.
\newblock In {\em Proceedings of the IEEE conference on computer vision and
  pattern recognition}, pages 3253--3261, 2018.

\bibitem{sedaghat2016orientation}
Nima Sedaghat, Mohammadreza Zolfaghari, Ehsan Amiri, and Thomas Brox.
\newblock Orientation-boosted voxel nets for 3d object recognition.
\newblock {\em arXiv preprint arXiv:1604.03351}, 2016.

\bibitem{su2015multi}
Hang Su, Subhransu Maji, Evangelos Kalogerakis, and Erik Learned-Miller.
\newblock Multi-view convolutional neural networks for 3d shape recognition.
\newblock In {\em Proceedings of the IEEE international conference on computer
  vision}, pages 945--953, 2015.

\bibitem{sun2019videobert}
Chen Sun, Austin Myers, Carl Vondrick, Kevin Murphy, and Cordelia Schmid.
\newblock Videobert: A joint model for video and language representation
  learning.
\newblock In {\em Proceedings of the IEEE/CVF International Conference on
  Computer Vision}, pages 7464--7473, 2019.

\bibitem{tan2019lxmert}
Hao Tan and Mohit Bansal.
\newblock Lxmert: Learning cross-modality encoder representations from
  transformers.
\newblock {\em arXiv preprint arXiv:1908.07490}, 2019.

\bibitem{uy2019revisiting}
Mikaela~Angelina Uy, Quang-Hieu Pham, Binh-Son Hua, Thanh Nguyen, and Sai-Kit
  Yeung.
\newblock Revisiting point cloud classification: A new benchmark dataset and
  classification model on real-world data.
\newblock In {\em Proceedings of the IEEE/CVF international conference on
  computer vision}, pages 1588--1597, 2019.

\bibitem{wang2023mvcontrast}
Luequan Wang, Hongbin Xu, and Wenxiong Kang.
\newblock Mvcontrast: Unsupervised pretraining for multi-view 3d object
  recognition.
\newblock {\em Machine Intelligence Research}, pages 1--12, 2023.

\bibitem{wang2019dynamic}
Yue Wang, Yongbin Sun, Ziwei Liu, Sanjay~E Sarma, Michael~M Bronstein, and
  Justin~M Solomon.
\newblock Dynamic graph cnn for learning on point clouds.
\newblock {\em Acm Transactions On Graphics (tog)}, 38(5):1--12, 2019.

\bibitem{wang2022p2p}
Ziyi Wang, Xumin Yu, Yongming Rao, Jie Zhou, and Jiwen Lu.
\newblock P2p: Tuning pre-trained image models for point cloud analysis with
  point-to-pixel prompting.
\newblock {\em arXiv preprint arXiv:2208.02812}, 2022.

\bibitem{wu20153d}
Zhirong Wu, Shuran Song, Aditya Khosla, Fisher Yu, Linguang Zhang, Xiaoou Tang,
  and Jianxiong Xiao.
\newblock 3d shapenets: A deep representation for volumetric shapes.
\newblock In {\em Proceedings of the IEEE conference on computer vision and
  pattern recognition}, pages 1912--1920, 2015.

\bibitem{xie2020pointcontrast}
Saining Xie, Jiatao Gu, Demi Guo, Charles~R Qi, Leonidas Guibas, and Or Litany.
\newblock Pointcontrast: Unsupervised pre-training for 3d point cloud
  understanding.
\newblock In {\em European conference on computer vision}, pages 574--591.
  Springer, 2020.

\bibitem{xu2022simple}
Mengde Xu, Zheng Zhang, Fangyun Wei, Yutong Lin, Yue Cao, Han Hu, and Xiang
  Bai.
\newblock A simple baseline for open-vocabulary semantic segmentation with
  pre-trained vision-language model.
\newblock In {\em European Conference on Computer Vision}, pages 736--753.
  Springer, 2022.

\bibitem{xu2019disn}
Qiangeng Xu, Weiyue Wang, Duygu Ceylan, Radomir Mech, and Ulrich Neumann.
\newblock Disn: Deep implicit surface network for high-quality single-view 3d
  reconstruction.
\newblock {\em Advances in Neural Information Processing Systems}, 32, 2019.

\bibitem{yao2021filip}
Lewei Yao, Runhui Huang, Lu Hou, Guansong Lu, Minzhe Niu, Hang Xu, Xiaodan
  Liang, Zhenguo Li, Xin Jiang, and Chunjing Xu.
\newblock Filip: Fine-grained interactive language-image pre-training.
\newblock {\em arXiv preprint arXiv:2111.07783}, 2021.

\bibitem{yao20223d}
Yuan Yao, Yuanhan Zhang, Zhenfei Yin, Jiebo Luo, Wanli Ouyang, and Xiaoshui
  Huang.
\newblock 3d point cloud pre-training with knowledge distillation from 2d
  images.
\newblock {\em arXiv preprint arXiv:2212.08974}, 2022.

\bibitem{you2019large}
Yang You, Jing Li, Sashank Reddi, Jonathan Hseu, Sanjiv Kumar, Srinadh
  Bhojanapalli, Xiaodan Song, James Demmel, Kurt Keutzer, and Cho-Jui Hsieh.
\newblock Large batch optimization for deep learning: Training bert in 76
  minutes.
\newblock {\em arXiv preprint arXiv:1904.00962}, 2019.

\bibitem{yu2022point}
Xumin Yu, Lulu Tang, Yongming Rao, Tiejun Huang, Jie Zhou, and Jiwen Lu.
\newblock Point-bert: Pre-training 3d point cloud transformers with masked
  point modeling.
\newblock In {\em Proceedings of the IEEE/CVF Conference on Computer Vision and
  Pattern Recognition}, pages 19313--19322, 2022.

\bibitem{zhai2019largescale}
Xiaohua Zhai, Joan Puigcerver, Alexander Kolesnikov, Pierre Ruyssen, Carlos
  Riquelme, Mario Lucic, Josip Djolonga, Andre~Susano Pinto, Maxim Neumann,
  Alexey Dosovitskiy, Lucas Beyer, Olivier Bachem, Michael Tschannen, Marcin
  Michalski, Olivier Bousquet, Sylvain Gelly, and Neil Houlsby.
\newblock A large-scale study of representation learning with the visual task
  adaptation benchmark, 2019.

\bibitem{zhang2021tip}
Renrui Zhang, Rongyao Fang, Peng Gao, Wei Zhang, Kunchang Li, Jifeng Dai, Yu
  Qiao, and Hongsheng Li.
\newblock Tip-adapter: Training-free clip-adapter for better vision-language
  modeling.
\newblock {\em arXiv preprint arXiv:2111.03930}, 2021.

\bibitem{zhang2022pointclip}
Renrui Zhang, Ziyu Guo, Wei Zhang, Kunchang Li, Xupeng Miao, Bin Cui, Yu Qiao,
  Peng Gao, and Hongsheng Li.
\newblock Pointclip: Point cloud understanding by clip.
\newblock In {\em Proceedings of the IEEE/CVF Conference on Computer Vision and
  Pattern Recognition}, pages 8552--8562, 2022.

\bibitem{zhang2022mining}
Zekang Zhang, Guangyu Gao, Zhiyuan Fang, Jianbo Jiao, and Yunchao Wei.
\newblock Mining unseen classes via regional objectness: A simple baseline for
  incremental segmentation.
\newblock {\em Advances in Neural Information Processing Systems},
  35:24340--24353, 2022.

\bibitem{zhao2021point}
Hengshuang Zhao, Li Jiang, Jiaya Jia, Philip~HS Torr, and Vladlen Koltun.
\newblock Point transformer.
\newblock In {\em Proceedings of the IEEE/CVF International Conference on
  Computer Vision}, pages 16259--16268, 2021.

\bibitem{zhou2017scene}
Bolei Zhou, Hang Zhao, Xavier Puig, Sanja Fidler, Adela Barriuso, and Antonio
  Torralba.
\newblock Scene parsing through ade20k dataset.
\newblock In {\em Proceedings of the IEEE conference on computer vision and
  pattern recognition}, pages 633--641, 2017.

\bibitem{zhou2021ibot}
Jinghao Zhou, Chen Wei, Huiyu Wang, Wei Shen, Cihang Xie, Alan Yuille, and Tao
  Kong.
\newblock ibot: Image bert pre-training with online tokenizer.
\newblock {\em arXiv preprint arXiv:2111.07832}, 2021.

\end{thebibliography}
}

\clearpage
\appendix
\section{Details of Loss Function}
$l_{intra}^i(\cdot)$ and $l_{cross}^i(\cdot)$ are InfoNCE-based loss. They are formulated as follows,
\begin{gather}
    s_{intra}^i(d_1,d_2) = \sum\limits_{k=1}^N e(\mathbf{F}_{i, d_1}^D, \mathbf{F}_{k, d_1}^D) + e(\mathbf{F}_{i, d_1}^D, \mathbf{F}_{k, d_2}^D), \\
    l_{intra}^i(d_1,d_2) = - \log \frac{e(\mathbf{F}_{i, d_1}^D, \mathbf{F}_{i, d_2}^D)}{s_{intra}^i(d_1,d_2) - e(\mathbf{F}_{i, d_1}^D, \mathbf{F}_{i, d_1}^D)}. \\
    s_{cross}^i(D, I) = \sum\limits_{k=1}^{N} e(\mathbf{F}_{i}^D, \mathbf{F}_{k}^D) + e(\mathbf{F}_{i}^D, \mathbf{F}_{k}^I), \\
    l_{cross}^i(D, I) = - \log \frac{e(\mathbf{F}_{i}^D, \mathbf{F}_{i}^I)}{s_{cross}^i(D, I) - e(\mathbf{F}_{i}^D, \mathbf{F}_{i}^D)}.
\end{gather}
Here, $e(a, b) = \exp(a \cdot b^T / \tau)$. We set the temperature coefficient $\tau = 0.7$.

\section{Complexity Analysis}
Based on the experiments on fully-supervised classification, we further report the computation costs for evaluation and the parameter sizes for training in Tab.~\ref{tab:complexity} to analyze the complexity of compared models.

Since PointCLIP achieves its best accuracy when using ResNet50x16, its computation cost is higher than our CLIP2Point with ViT-B/32.
While the computation cost of P2P is even higher, its cost needs to be multiplied by 40 (the number of views) as P2P infers a single view at one time.
Our CLIP2Point achieves a higher accuracy than P2P (HorNet-L) with a much lower computation cost.

On the other hand, embedded with GDPA module, CLIP2Point contains fewer training parameters than those full-tuning methods. Only by tuning lightweight adapters, CLIP2Point can outperform the state-of-the-art pre-training method Point-MAE.

\textbf{Note that}, the low computation cost of Transformer is because the grouping and gathering mechanisms for point cloud are not included in the calculation of MACs. Thus, comparing the MACs values with 3D networks is not fair.

\begin{table}[!htbp]
    \centering
    \caption{The computation costs for evaluation and the parameter sizes for training, based on fully-supervised classification.}
    \begin{tabular}{l|cc}
        \hline
        Methods & Eval.~MACs(G) & Tr.~Param.(M) \\
        \hline
        MVCNN & 43.72 & 11.20 \\
        SimpleView & 53.38 & 12.76 \\
        MVTN & 45.97 & 27.06 \\
        PointCLIP & 227.42 & 5.51 \\
        P2P: ResNet-101 & 11.96$(\times40)$ & 0.25 \\
        P2P: ConvNeXt-L & 38.51$(\times40)$ & 0.14 \\
        P2P: HorNet-L & 38.72$(\times40)$ & 1.01 \\
        \hline
        Transformer & 2.40 & 22.10 \\
        + Point-BERT & 2.40 & 22.10 \\
        + Point-MAE & 2.40 & 22.10 \\
        \hline
        CLIP2Point (Ours) & 88.23 & 5.78 \\
        \hline
    \end{tabular}
    \label{tab:complexity}
\end{table}

\section{Application on Scene-Level Tasks}
Analogous to CLIP, we design our pre-training pipeline in an instance-level paradigm. We note that the knowledge in CLIP is more suitable to be used in image classification and retrieval, and it is also hard to directly leverage fine-grained knowledge from CLIP.
Thus, existing works~\cite{xu2022simple,lu2022open} usually adopt extra modules to provide possible proposals, and CLIP still works as a classifier. Following~\cite{lu2022open}, we conduct open-vocabulary 3D detection experiments, using our CLIP2Point to classify the bounding box generated by 3D detectors. In Tab.~\ref{tab:detect}, CLIP2Point outperforms two 3D detection networks and two cross-modal methods based on pre-training, indicating that \textbf{CLIP2Point can adapt to open-world scene-level tasks}. In contrast to OV-3DETIC specifically trained on 3D detection datasets with the distillation of a 2D detector, our mAP is relatively low, owing to noised point cloud data in proposed bounding boxes. 
Nonetheless, the result verifies the feasibility of such knowledge transfer in scene-level tasks.
%
In the conclusion of the main text, we also mention that real-world training data can further enhance tasks related to real scenes in future work. Due to that CLIP2Point is effective in classifying objects, it can be naturally applied to scene-level tasks with proper proposals.

\begin{table}
  \centering
  \caption{Open-Vocabulary 3D Detection on ScanNet.}
  \begin{tabular}{ccc}
    \hline
    Method & Training Data & $mAP_{25}$ \\
    \hline
    VoteNet & \multicolumn{1}{c}{\multirow{2}{*}{Seen(Train)-Unseen(Test)}} & 0.04 \\
    3DETR & \multicolumn{1}{c}{} & 1.11 \\
    \hline
    Image2Point & \multicolumn{1}{c}{\multirow{3}{*}{Zero-Shot}} & 0.84 \\
    PointCLIP & \multicolumn{1}{c}{} & 3.09 \\
    CLIP2Point & \multicolumn{1}{c}{} & 3.71 \\
    \hline
    OV-3DETIC & 2D\&3D Detection Data & 12.65 \\
    \hline
  \end{tabular}
  \label{tab:detect}
\end{table}



\section{Why Not Applying CLIP to 3D Networks?}

Since 3D backbones can be applied to downstream tasks more easily, a natural idea is whether CLIP pre-training knowledge can be directly applied to 3D networks. In fact, previous works~\cite{afham2022crosspoint,li2022closer} have already demonstrated the effectiveness of such 2D-3D transfer. However, these methods achieve a sound result only after well fine-tuning on specific downstream datasets (\textit{e.g.}, DGCNN pre-trained by CrossPoint even cannot surpass PointCLIP in the few-shot experiment in our main text), which means they can hardly adapt to 3D tasks with 2D knowledge alone. We replace the depth encoder in CLIP2Point with a 3D encoder Point Transformer~\cite{zhao2021point}, while getting a \textbf{20.83\%} accuracy in ModelNet40 zero-shot classification after pre-training. We summarize two reasons for the bad result: 1) Features extracted by 2D and 3D encoders have different granularities: 2D encoders extract single-view features, while 3D encoders can aggregate a complete 3D object; 2) The gap between the parameter sizes of 2D and 3D encoders is large (\textit{e.g.}, ViT-B/32 in CLIP contains 87.85M parameters, but DGCNN only contains 0.98M parameters): 2D pre-training knowledge cannot completely transfer to small 3D encoders.

Nonetheless, it is still promising future work to directly transfer CLIP knowledge to 3D networks.

\section{Rendering Details}
\begin{figure*}[!htbp]
    \centering
    \begin{tabular}{ccccc}
        \includegraphics[width=0.17\textwidth]{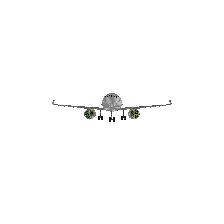} &
        \includegraphics[width=0.17\textwidth]{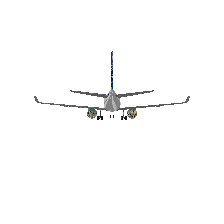} &
        \includegraphics[width=0.17\textwidth]{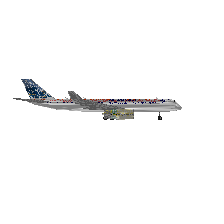} &
        \includegraphics[width=0.17\textwidth]{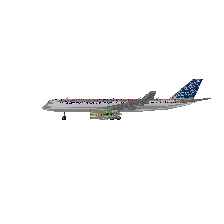} &
        \includegraphics[width=0.17\textwidth]{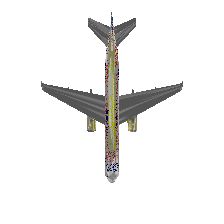} \\
        front & back & left & right & top \\
        \includegraphics[width=0.17\textwidth]{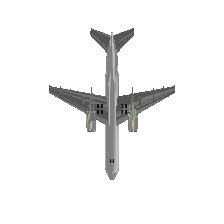} &
        \includegraphics[width=0.17\textwidth]{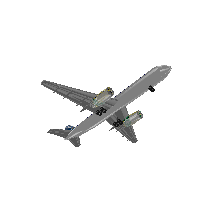} &
        \includegraphics[width=0.17\textwidth]{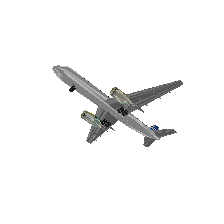} &
        \includegraphics[width=0.17\textwidth]{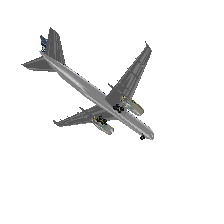} &
        \includegraphics[width=0.17\textwidth]{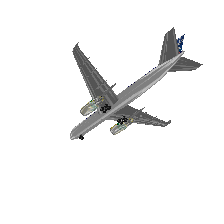} \\
        bottom & front-left & front-right & back-left & back-right \\
    \end{tabular}
    \caption{Visualization of multi-view RGB images for an airplane.}
    \label{fig:10view}
\end{figure*}

Following MVTN~\cite{hamdi2021mvtn}, we render 3D models to RGB images with Pytorch3D~\cite{lassner2020pulsar}. We first load mesh objects with texture information from ShapeNetCore v2. We choose 10 views in a spherical configuration, and then use \textbf{MeshRasterizer} and \textbf{HardPhongShader} in Pytorch3D.render, with the colors of backgrounds and lights both white. For zero-shot evaluation, we use 6 orthogonal views: front, back, left, right, top, and bottom. We add four corner views for pre-training and downstream learning. The view distance is initialized as 1, and the random range of distance in pre-training is $[0.9, 1.1)$. We visualize ten views of an airplane in Fig.~\ref{fig:10view}.

\section{Dataset Visualization}
\begin{figure*}[!htbp]
    \centering
    \begin{tabular}{ccccc}
        \includegraphics[width=0.08\textwidth]{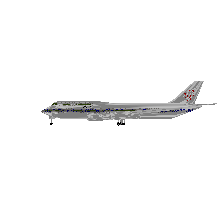} &
        \includegraphics[width=0.08\textwidth]{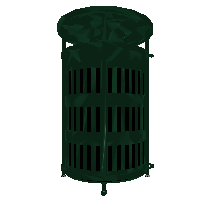} &
        \includegraphics[width=0.08\textwidth]{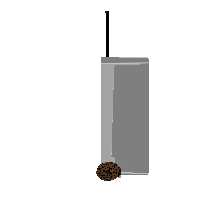} &
        \includegraphics[width=0.08\textwidth]{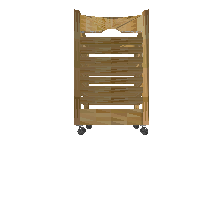} &
        \includegraphics[width=0.08\textwidth]{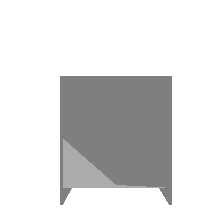} \\
        airplane-image & ashcan-image & bag-image & basket-image & bathtub-image \\
        
        \includegraphics[width=0.08\textwidth]{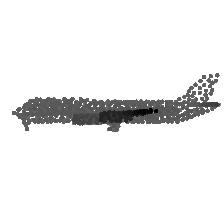} &
        \includegraphics[width=0.08\textwidth]{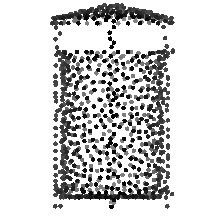} &
        \includegraphics[width=0.08\textwidth]{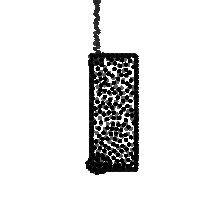} &
        \includegraphics[width=0.08\textwidth]{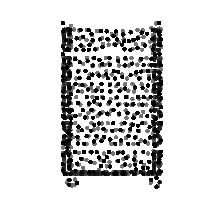} &
        \includegraphics[width=0.08\textwidth]{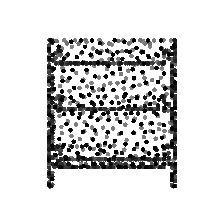} \\
        \vspace{2em}
        airplane-depth & ashcan-depth & bag-depth & basket-depth & bathtub-depth\\
        \includegraphics[width=0.08\textwidth]{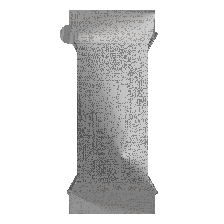} &
        \includegraphics[width=0.08\textwidth]{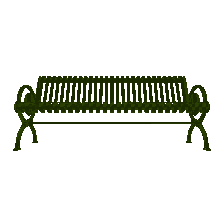} &
        \includegraphics[width=0.08\textwidth]{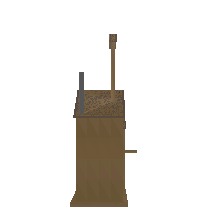} &
        \includegraphics[width=0.08\textwidth]{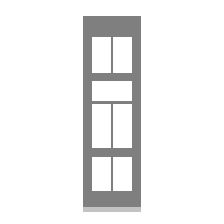} &
        \includegraphics[width=0.08\textwidth]{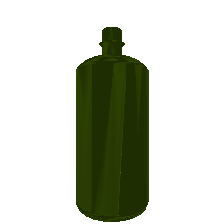} \\
        bed-image & bench-image & birdhouse-image & bookshelf-image & bottle-image \\
        \includegraphics[width=0.08\textwidth]{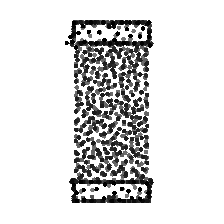} &
        \includegraphics[width=0.08\textwidth]{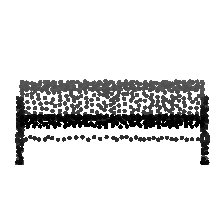} &
        \includegraphics[width=0.08\textwidth]{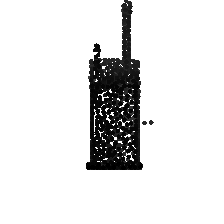} &
        \includegraphics[width=0.08\textwidth]{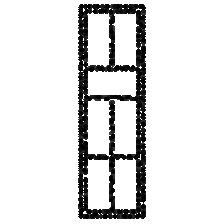} &
        \includegraphics[width=0.08\textwidth]{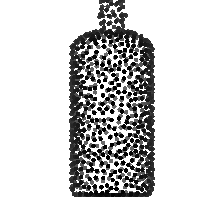} \\
        \vspace{2em}
        bed-depth & bench-depth & birdhouse-depth & bookshelf-depth & bottle-depth \\
        
        \includegraphics[width=0.08\textwidth]{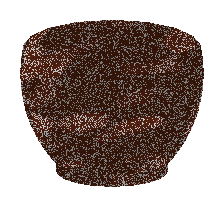} &
        \includegraphics[width=0.08\textwidth]{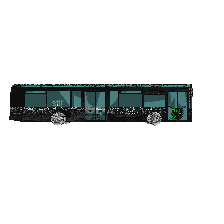} &
        \includegraphics[width=0.08\textwidth]{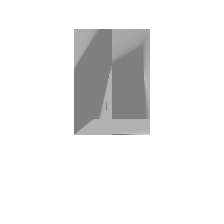} &
        \includegraphics[width=0.08\textwidth]{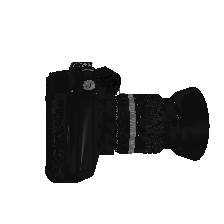} &
        \includegraphics[width=0.08\textwidth]{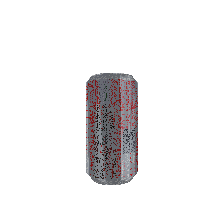} \\
        bowl-image & bus-image & cabinet-image & camera-image & can-image \\
        \includegraphics[width=0.08\textwidth]{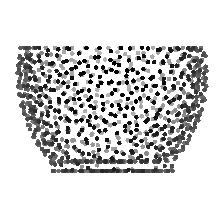} &
        \includegraphics[width=0.08\textwidth]{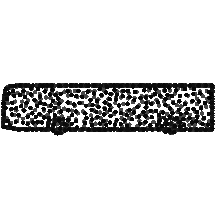} &
        \includegraphics[width=0.08\textwidth]{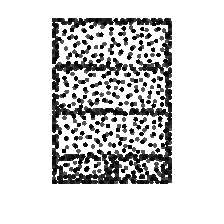} &
        \includegraphics[width=0.08\textwidth]{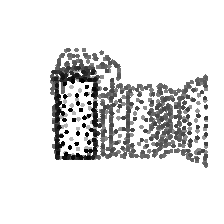} &
        \includegraphics[width=0.08\textwidth]{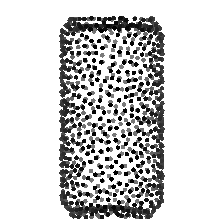} \\
        \vspace{2em}
        bowl-depth & bus-depth & cabinet-depth & camera-depth & can-depth \\
        
        \includegraphics[width=0.08\textwidth]{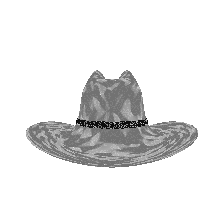} &
        \includegraphics[width=0.08\textwidth]{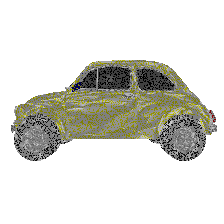} &
        \includegraphics[width=0.08\textwidth]{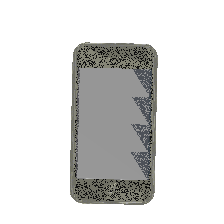} &
        \includegraphics[width=0.08\textwidth]{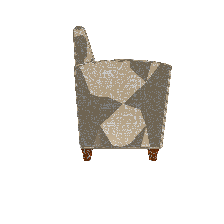} &
        \includegraphics[width=0.08\textwidth]{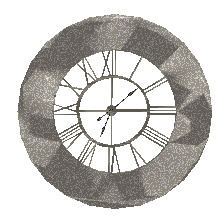} \\
        cap-image & car-image & phone-image & chair-image & clock-image \\
        \includegraphics[width=0.08\textwidth]{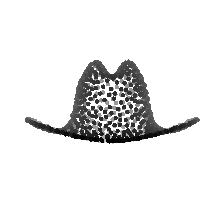} &
        \includegraphics[width=0.08\textwidth]{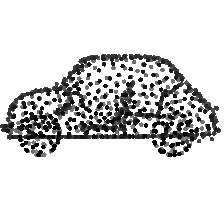} &
        \includegraphics[width=0.08\textwidth]{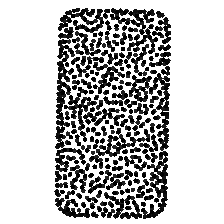} &
        \includegraphics[width=0.08\textwidth]{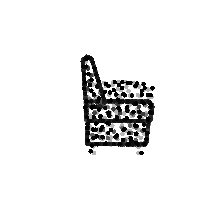} &
        \includegraphics[width=0.08\textwidth]{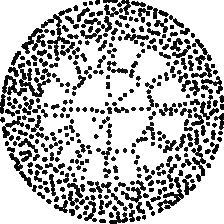} \\
        cap-depth & car-depth & phone-depth & chair-depth & clock-depth \\
    \end{tabular}
    \caption{Rendered RGB images of Category 1 $\sim$ Category 20 on ShapeNet.}
    \label{fig:rendered1}
\end{figure*}

\begin{figure*}[!htbp]
    \centering
    \begin{tabular}{ccccc}
        \includegraphics[width=0.08\textwidth]{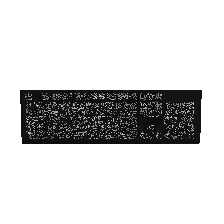} &
        \includegraphics[width=0.08\textwidth]{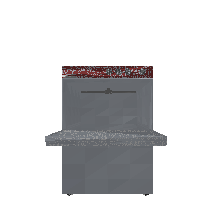} &
        \includegraphics[width=0.08\textwidth]{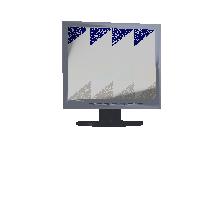} &
        \includegraphics[width=0.08\textwidth]{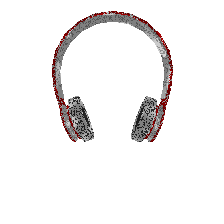} &
        \includegraphics[width=0.08\textwidth]{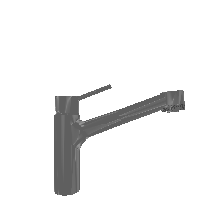} \\
        keyboard-image & dishwasher-image & display-image & earphone-image & faucet-image \\
        \includegraphics[width=0.08\textwidth]{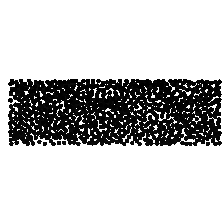} &
        \includegraphics[width=0.08\textwidth]{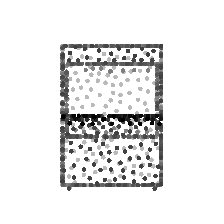} &
        \includegraphics[width=0.08\textwidth]{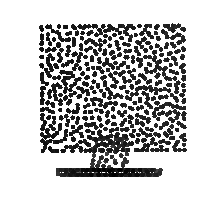} &
        \includegraphics[width=0.08\textwidth]{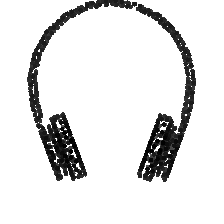} &
        \includegraphics[width=0.08\textwidth]{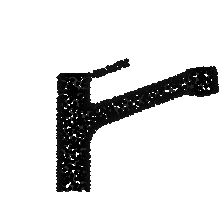} \\
        \vspace{2em}
        keyboard-depth & dishwasher-depth & display-depth & earphone-depth & faucet-depth \\
        
        \includegraphics[width=0.08\textwidth]{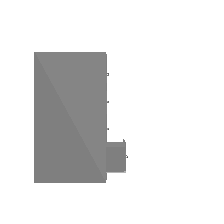} &
        \includegraphics[width=0.08\textwidth]{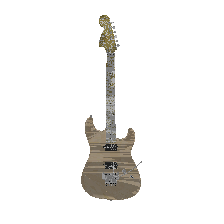} &
        \includegraphics[width=0.08\textwidth]{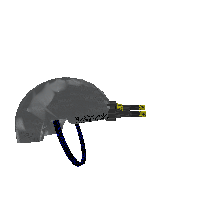} &
        \includegraphics[width=0.08\textwidth]{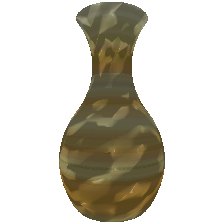} &
        \includegraphics[width=0.08\textwidth]{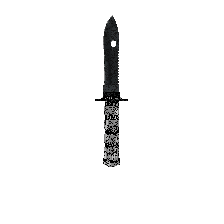} \\
        file-image & guitar-image & helmet-image & jar-image & knife-image \\
        \includegraphics[width=0.08\textwidth]{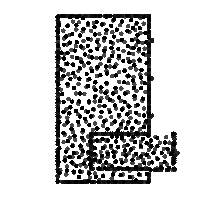} &
        \includegraphics[width=0.08\textwidth]{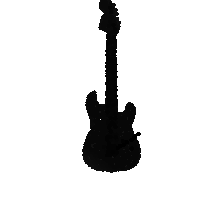} &
        \includegraphics[width=0.08\textwidth]{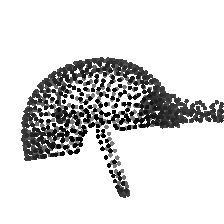} &
        \includegraphics[width=0.08\textwidth]{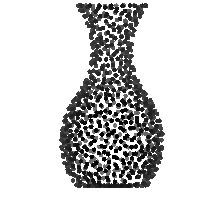} &
        \includegraphics[width=0.08\textwidth]{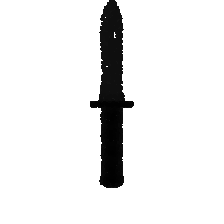} \\
        \vspace{2em}
        file-depth & guitar-depth & helmet-depth & jar-depth & knife-depth \\
        
        \includegraphics[width=0.08\textwidth]{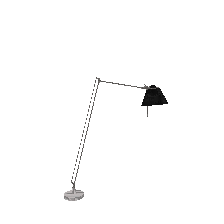} &
        \includegraphics[width=0.08\textwidth]{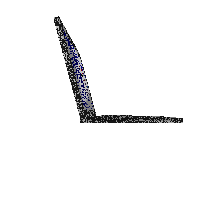} &
        \includegraphics[width=0.08\textwidth]{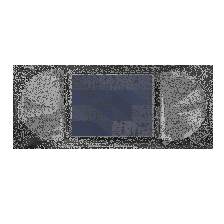} &
        \includegraphics[width=0.08\textwidth]{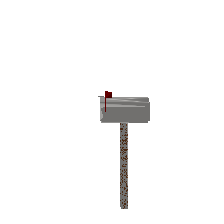} &
        \includegraphics[width=0.08\textwidth]{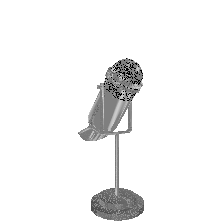} \\
        lamp-image & laptop-image & loudspeaker-image & mailbox-image & microphone-image \\
        \includegraphics[width=0.08\textwidth]{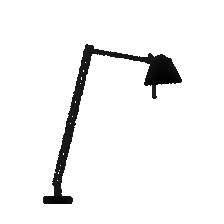} &
        \includegraphics[width=0.08\textwidth]{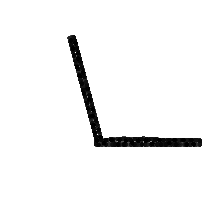} &
        \includegraphics[width=0.08\textwidth]{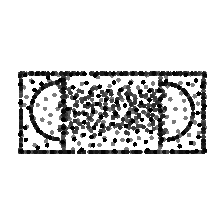} &
        \includegraphics[width=0.08\textwidth]{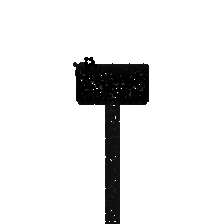} &
        \includegraphics[width=0.08\textwidth]{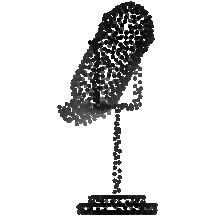} \\
        \vspace{2em}
        lamp-depth & laptop-depth & loudspeaker-depth & mailbox-depth & microphone-depth \\
        
        \includegraphics[width=0.08\textwidth]{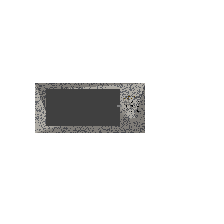} &
        \includegraphics[width=0.08\textwidth]{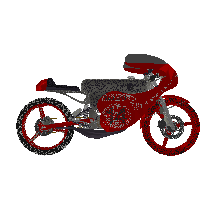} &
        \includegraphics[width=0.08\textwidth]{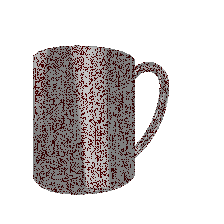} &
        \includegraphics[width=0.08\textwidth]{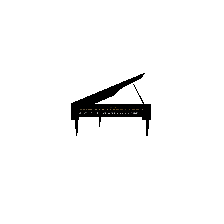} &
        \includegraphics[width=0.08\textwidth]{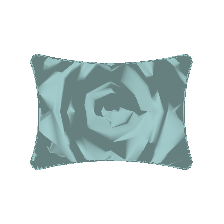} \\
        microwave-image & motorcycle-image & mug-image & piano-image & pillow-image \\
        \includegraphics[width=0.08\textwidth]{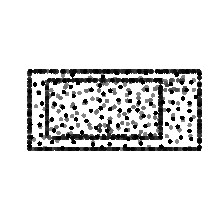} &
        \includegraphics[width=0.08\textwidth]{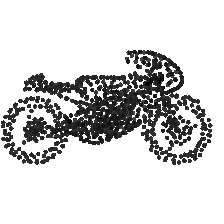} &
        \includegraphics[width=0.08\textwidth]{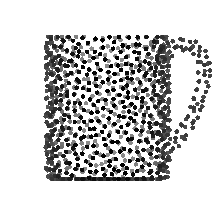} &
        \includegraphics[width=0.08\textwidth]{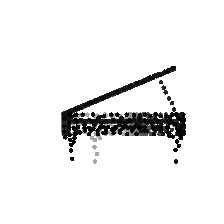} &
        \includegraphics[width=0.08\textwidth]{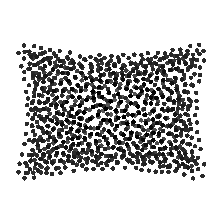} \\
        microwave-depth & motorcycle-depth & mug-depth & piano-depth & pillow-depth \\
    \end{tabular}
    \caption{Rendered RGB images of Category 21 $\sim$ Category 40 on ShapeNet.}
    \label{fig:rendered2}
\end{figure*}

\begin{figure*}[!htbp]
    \centering
    \begin{tabular}{ccccc}
        \includegraphics[width=0.08\textwidth]{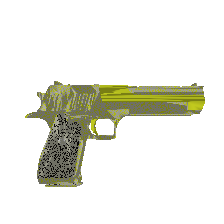} &
        \includegraphics[width=0.08\textwidth]{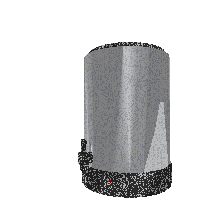} &
        \includegraphics[width=0.08\textwidth]{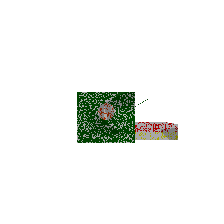} &
        \includegraphics[width=0.08\textwidth]{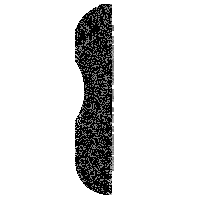} &
        \includegraphics[width=0.08\textwidth]{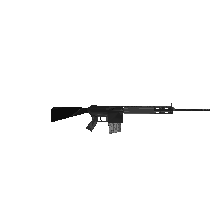} \\
        pistol-image & pot-image & printer-image & control-image & rifle-image \\
        \includegraphics[width=0.08\textwidth]{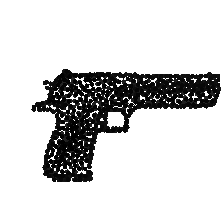} &
        \includegraphics[width=0.08\textwidth]{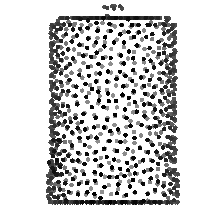} &
        \includegraphics[width=0.08\textwidth]{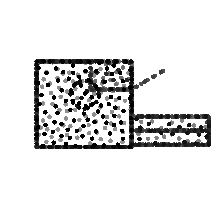} &
        \includegraphics[width=0.08\textwidth]{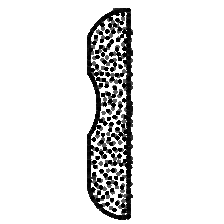} &
        \includegraphics[width=0.08\textwidth]{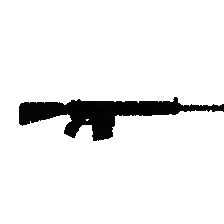} \\
        \vspace{2em}
        pistol-depth & pot-depth & printer-depth & control-depth & rifle-depth \\
        
        \includegraphics[width=0.08\textwidth]{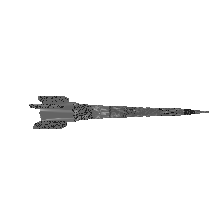} &
        \includegraphics[width=0.08\textwidth]{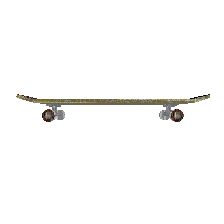} &
        \includegraphics[width=0.08\textwidth]{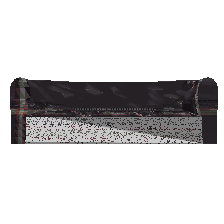} &
        \includegraphics[width=0.08\textwidth]{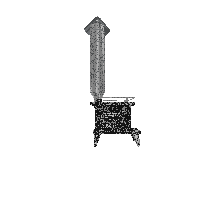} &
        \includegraphics[width=0.08\textwidth]{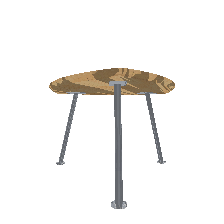} \\
        rocket-image & skateboard-image & sofa-image & stove-image & table-image \\
        \includegraphics[width=0.08\textwidth]{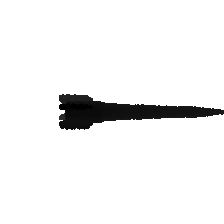} &
        \includegraphics[width=0.08\textwidth]{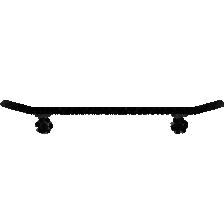} &
        \includegraphics[width=0.08\textwidth]{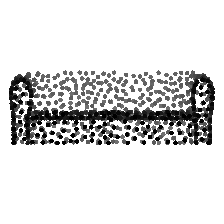} &
        \includegraphics[width=0.08\textwidth]{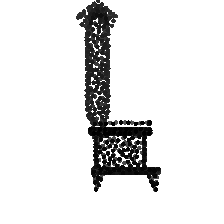} &
        \includegraphics[width=0.08\textwidth]{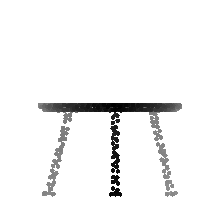} \\
        \vspace{2em}
        rocket-depth & skateboard-depth & sofa-depth & stove-depth & table-depth \\
        
        \includegraphics[width=0.08\textwidth]{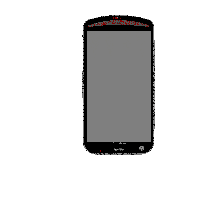} &
        \includegraphics[width=0.08\textwidth]{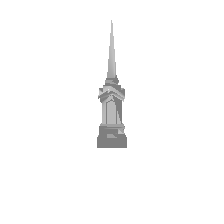} &
        \includegraphics[width=0.08\textwidth]{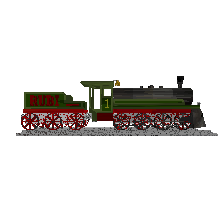} &
        \includegraphics[width=0.08\textwidth]{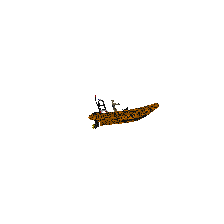} &
        \includegraphics[width=0.08\textwidth]{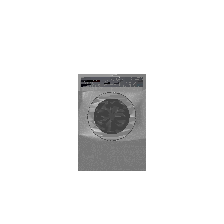} \\
        telephone-image & tower-image & train-image & vessel-image & washer-image \\
        \includegraphics[width=0.08\textwidth]{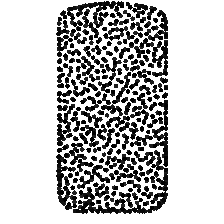} &
        \includegraphics[width=0.08\textwidth]{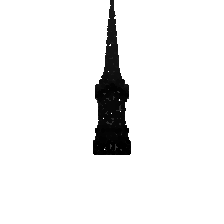} &
        \includegraphics[width=0.08\textwidth]{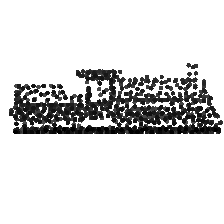} &
        \includegraphics[width=0.08\textwidth]{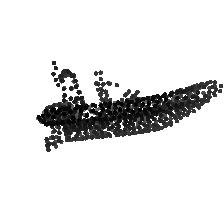} &
        \includegraphics[width=0.08\textwidth]{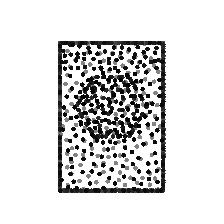} \\
        telephone-depth & tower-depth & train-depth & vessel-depth & washer-depth \\
    \end{tabular}
    \caption{Rendered RGB images of Category 41 $\sim$ Category 55 on ShapeNet.}
    \label{fig:rendered3}
\end{figure*}

We provide more visualization results in Fig.~\ref{fig:rendered1}, \ref{fig:rendered2}, \ref{fig:rendered3}. For each category in ShapeNet, we have a rendered RGB image and a corresponding depth map.

\end{document}